\PassOptionsToPackage{unicode}{hyperref}
\PassOptionsToPackage{hyphens}{url}
\PassOptionsToPackage{dvipsnames,svgnames,x11names}{xcolor}
\documentclass[
  12pt]{article}

\usepackage[utf8]{inputenc} 
\usepackage[T1]{fontenc}    
\usepackage{hyperref}       
\usepackage{url}            
\usepackage{booktabs}       
\usepackage{amsfonts}       
\usepackage{nicefrac}       
\usepackage{microtype}      
\usepackage{xcolor}         
\usepackage{amssymb,amsmath,amsthm,yhmath,graphicx,bm,bbm,enumitem,natbib,tikz,stackrel}
\usepackage{epsfig}

\usepackage{algorithm}
\usepackage{algpseudocode}
\usetikzlibrary{positioning,matrix,backgrounds,arrows,automata, calc}
\usetikzlibrary{arrows.meta, shapes, fit}

\usepackage[title]{appendix}
\usepackage{lipsum}

\usepackage{iftex}
\ifPDFTeX
  \usepackage[T1]{fontenc}
  \usepackage[utf8]{inputenc}
  \usepackage{textcomp} 
\else 
  \usepackage{unicode-math}
  \defaultfontfeatures{Scale=MatchLowercase}
  \defaultfontfeatures[\rmfamily]{Ligatures=TeX,Scale=1}
\fi
\usepackage{lmodern}
\ifPDFTeX\else  
\fi
\IfFileExists{upquote.sty}{\usepackage{upquote}}{}
\IfFileExists{microtype.sty}{
  \usepackage[]{microtype}
  \UseMicrotypeSet[protrusion]{basicmath} 
}{}
\makeatletter
\@ifundefined{KOMAClassName}{
  \IfFileExists{parskip.sty}{%
    \usepackage{parskip}
  }{
    \setlength{\parindent}{0pt}
    \setlength{\parskip}{6pt plus 2pt minus 1pt}}
}{
  \KOMAoptions{parskip=half}}
\makeatother
\usepackage{xcolor}
\makeatletter
\ifx\paragraph\undefined\else
  \let\oldparagraph\paragraph
  \renewcommand{\paragraph}{
    \@ifstar
      \xxxParagraphStar
      \xxxParagraphNoStar
  }
  \newcommand{\xxxParagraphStar}[1]{\oldparagraph*{#1}\mbox{}}
  \newcommand{\xxxParagraphNoStar}[1]{\oldparagraph{#1}\mbox{}}
\fi
\ifx\subparagraph\undefined\else
  \let\oldsubparagraph\subparagraph
  \renewcommand{\subparagraph}{
    \@ifstar
      \xxxSubParagraphStar
      \xxxSubParagraphNoStar
  }
  \newcommand{\xxxSubParagraphStar}[1]{\oldsubparagraph*{#1}\mbox{}}
  \newcommand{\xxxSubParagraphNoStar}[1]{\oldsubparagraph{#1}\mbox{}}
\fi
\makeatother

\usepackage{longtable,booktabs,array}
\usepackage{calc} 
\usepackage{etoolbox}
\makeatletter
\patchcmd\longtable{\par}{\if@noskipsec\mbox{}\fi\par}{}{}
\makeatother
\IfFileExists{footnotehyper.sty}{\usepackage{footnotehyper}}{\usepackage{footnote}}
\makesavenoteenv{longtable}
\usepackage{graphicx}
\makeatletter
\def\maxwidth{\ifdim\Gin@nat@width>\linewidth\linewidth\else\Gin@nat@width\fi}
\def\maxheight{\ifdim\Gin@nat@height>\textheight\textheight\else\Gin@nat@height\fi}
\makeatother
\setkeys{Gin}{width=\maxwidth,height=\maxheight,keepaspectratio}
\makeatletter
\def\fps@figure{htbp}
\makeatother

\makeatletter
\@ifpackageloaded{caption}{}{\usepackage{caption}}
\AtBeginDocument{%
\ifdefined\contentsname
  \renewcommand*\contentsname{Table of contents}
\else
  \newcommand\contentsname{Table of contents}
\fi
\ifdefined\listfigurename
  \renewcommand*\listfigurename{List of Figures}
\else
  \newcommand\listfigurename{List of Figures}
\fi
\ifdefined\listtablename
  \renewcommand*\listtablename{List of Tables}
\else
  \newcommand\listtablename{List of Tables}
\fi
\ifdefined\figurename
  \renewcommand*\figurename{Figure}
\else
  \newcommand\figurename{Figure}
\fi
\ifdefined\tablename
  \renewcommand*\tablename{Table}
\else
  \newcommand\tablename{Table}
\fi
}
\@ifpackageloaded{float}{}{\usepackage{float}}
\floatstyle{ruled}
\@ifundefined{c@chapter}{\newfloat{codelisting}{h}{lop}}{\newfloat{codelisting}{h}{lop}[chapter]}
\floatname{codelisting}{Listing}

\makeatother
\makeatletter
\@ifpackageloaded{caption}{}{\usepackage{caption}}
\@ifpackageloaded{subcaption}{}{\usepackage{subcaption}}
\makeatother

\ifLuaTeX
  \usepackage{selnolig}  
\fi
\usepackage[]{natbib}
\usepackage{bookmark}

\IfFileExists{xurl.sty}{\usepackage{xurl}}{} 
\hypersetup{
  pdftitle={Title},
  pdfauthor={Author 1; Author 2},
  pdfkeywords={3 to 6 keywords, that do not appear in the title},
  colorlinks=true,
  linkcolor={blue},
  filecolor={Maroon},
  citecolor={Blue},
  urlcolor={Blue},
  pdfcreator={LaTeX via pandoc}}

\newcommand{\anon}{1}

\newcommand{\R}{\mathbb{R}}
\renewcommand{\b }{\mathbf}

\def\wh{\widehat}
\def\wt{\widetilde}
\def\trans{^{\top}}
\def\bse{\begin{eqnarray*}}
\def\ese{\end{eqnarray*}}
\def\be{\begin{eqnarray}}
\def\ee{\end{eqnarray}}
\def\bb{\mbox{\boldmath$\beta$}}
\def\beps{{\mbox{\boldmath$\epsilon$}}}
\def\boxit#1{\vbox{\hrule\hbox{\vrule\kern6pt
\vbox{\kern6pt#1\kern6pt}\kern6pt\vrule}\hrule}}
\theoremstyle{definition}
\newtheorem{definition}{Definition}[section]
\theoremstyle{remark}
\newtheorem*{remark}{Remark}
\newtheorem{theorem}{Theorem}

\def\bb{\mbox{\boldmath$\beta$}}

\def\btheta{\mbox{\boldmath$\theta$}}
\def\beps{{\mbox{\boldmath$\epsilon$}}}

\def\g{{\bf g}}

\def\h{{\bf h}}

\def\t{{\bf t}}

\def\s{{\bf s}}

\def\v{{\bf v}}

\def\X{{\bf X}}

\def\x{{\bf x}}

\def\Y{{\bf Y}}

\def\diag{\hbox{diag}}

\def\boxit#1{\vbox{\hrule\hbox{\vrule\kern6pt
\vbox{\kern6pt#1\kern6pt}\kern6pt\vrule}\hrule}}

\def\var{\hbox{var}}

\def\g{{\bf g}}

\begin{document}

\def\spacingset#1{\renewcommand{\baselinestretch}%
{#1}\small\normalsize} \spacingset{1}


\if1\anon
{
  \title{\bf Distributional Deep Learning for Super-Resolution of 4D Flow MRI under Domain Shift}
  \author{Xiaoyi Wen
   \hspace{.2cm}\\
    and \\
    Fei Jiang \\
    Department of Epidemiology and Biostatistics, \\
    University of California, San Francisco}
  \maketitle
} \fi

\if0\anon
{
  \bigskip
  \bigskip
  \bigskip
  \begin{center}
    {\LARGE\bf Distributional Deep Learning for Super-Resolution of 4D Flow MRI under Domain Shift}
\end{center}
  \medskip
} \fi

\bigskip
\begin{abstract}
Super-resolution is widely used in medical imaging to enhance low-quality data, reducing scan time and improving abnormality detection.  Conventional super-resolution approaches typically rely on paired datasets of downsampled and original high resolution images, training models to reconstruct high resolution images from their artificially degraded counterparts. However, in real-world clinical settings, low resolution data often arise from acquisition mechanisms that differ significantly from simple downsampling. As a result, these inputs may lie outside the domain of the training data, leading to poor model generalization due to domain shift. To address this limitation, we propose a distributional deep learning framework that improves model robustness and domain generalization. We develop this approch for enhancing the resolution of 4D Flow MRI (4DF). This is a novel imaging modality that captures hemodynamic flow velocity and clinically relevant metrics such as vessel wall stress. These metrics are critical for assessing aneurysm rupture risk. Our model is initially trained on high resolution computational fluid dynamics (CFD) simulations and their downsampled counterparts. It is then fine-tuned on a small, harmonized dataset of paired 4D Flow MRI and CFD samples. We derive the theoretical properties of our distributional estimators and demonstrate that our framework significantly outperforms traditional deep learning approaches through real data applications. This highlights the effectiveness of distributional learning in addressing domain shift and improving super-resolution performance in clinically realistic scenarios.
\end{abstract}

\noindent%
\vfill

\newpage
\spacingset{1.8} 

\section{Introduction}

Cerebral aneurysms, prevalent in approximately 6\% of the general population, pose significant clinical challenges due to their potential for rupture, leading to high morbidity and mortality. Fortunately, most unruptured intracranial aneurysms (UIAs) rarely cause symptoms and do not require an invasive treatment that may itself causes severe cerebrovascular disorders. However, it is very difficult to predict which UIAs will rupture. Recent evaluations of the hemodynamic features of UIAs, using 4D Flow MRI (4DF), have shown promising results that suggest specific hemodynamic variables may have a great impact on aneurysm growth or rupture \citep{hope2010evaluation, futami2016inflow, ferdian20204dflownet}.

4DF offers direct in vivo blood flow measurements. It captures the actual hemodynamics during the scan, so it inherently includes the true physiological conditions, which makes 4DF highly valuable for patient-specific assessment. However, the quality of the 4DF data is constrained by measurement noise and imaging artifacts \citep{rutkowski2021enhancement}. In addition,  the spatial resolution of 4DF data is limited due to compromises made to ensure patient comfort during scanning. As a result, small-scale flow characteristics, especially those near vessel walls, are often overlooked \citep{callaghan2017spatial}, limiting  the accuracy of  estimating the critical metrics such as wall shear stress and the shear concentration index \citep{cebral2011quantitative}.   In contrast, computational fluid dynamics (CFD) has been the
predominant methodology used to study hemodynamics in cerebral aneurysms. By applying the Navier–Stokes equations to patient-derived vascular geometry, CFD enables the simulation of high resolution, noise-free velocity fields that can be used to derive flow variables of interest \citep{karabasov2009compact}. However, CFD is a simulation-based method that strongly depends on the accuracy of modeling assumptions and the initial conditions applied to the Navier–Stokes equations. Therefore, using CFD in clinical practice is less feasiable given that there is open lack of expertise to process the simulation and outcomes are highly sensitive to patient-specific conditions and require careful validation. 

These differences highlight a strong motivation to leverage carefully modeled CFD data to enhance 4DF image resolution. A widely adopted approach involves training super-resolution  deep learning models on synthetic 4DF data derived from CFD simulations \citep{ferdian20204dflownet, ferdian2023cerebrovascular, long2023super, shone2023deep}. Beyond deep learning-based frameworks, \cite{perez2020towards} leverage proper orthogonal decomposition (POD) to extract dominant flow features from patient-specific CFD data, subsequently employing generalized dynamic mode decomposition (DMD) for 4DF super-resolution. Additionally, flow data assimilation methods employing Kalman filtering \citep{habibi2021integrating, gaidzik2021hemodynamic} have also demonstrated efficiency in achieving 4DF super-resolution.

Due to challenges in data harmonization,  paired 4DF and CFD datasets are limited. Consequently, most super-resolution methods are developed exclusively using CFD data, assuming that downsampled CFD can approximate the distribution of 4DF data. Models trained on downsampled and original CFD pairs are then directly applied to upsample 4DF data. While these approaches have demonstrated promising results on synthetic 4DF datasets, their performance on real 4DF data remains largely underexplored. A critical yet often overlooked issue is the domain discrepancy between downsampled CFD and actual 4DF data. For example, downsampled CFD still follows the physical principles such as mass conservation, whereas real 4DF data often violates these constraints. Moreover, variations in acquisition protocols and physiological conditions further exacerbate the distributional gap between the two domains \citep{cherry2022impact, black2023calibration}. This domain mismatch can significantly impair the performance of super-resolution models when using on clinical 4DF data \citep{shit2022srflow}.

In this paper, we propose a Distributional Super-Resolution (DSR) model aimed at enhancing the resolution of 4DF data. The DSR model demonstrates strong domain extrapolation capabilities, effectively handling scenarios where the input domain in the testing data deviates from that in the training data. Specifically, we employ a pre-additive model to capture the relationship between low-resolution ($\b X$) and high-resolution ($\b Y$) data through $\b Y = \g \left\{ \bb^\top \left( \b X + \beps \right)  \right\}$, where $\beps$ is a noise component. In our motivating dataset, the response $\b Y$ corresponds to high-resolution CFD measurements, while the predictors $\b X$ are the  low-resolution 4DF data. By incorporating noise into the training data, we effectively expand the domain of the training distribution. Our main goal is to learn the composit function $\g \circ \bb\trans$  from CFD data, and then apply the trained model to enhance the resolution of real-world 4DF data.
 
Building on the DSR model, we present the first comprehensive 4DF super-resolution framework validated on real intracranial 4DF data. We propose a local, patch-based super-resolution strategy specifically designed to enhance resolution on complex, irregular three-dimensional vascular geometries. This patch-based strategy ensures broad applicability across diverse structures. In addition, we implement a pre-training and fine-tuning paradigm that achieves superior performance even with limited training samples. We comprehensively evaluate the impact of each step in the paradigm on the final super-resolution results and justify the necessity of each step for the application.  To facilitate adoption, we provide a software package \hyperlink{https://github.com/wenxy123/DSR.git}{DSR} that can be readily used in other research settings. Together, this framework establishes a robust foundation for advancing 4DF super-resolution.

Our frameowrk is related to but distinct from existing super-resolution methods in several important aspects. First, most super-resolution techniques \citep{liang2021swinir, georgescu2023multimodal, ferdian20204dflownet} focus on learning the mapping between low- and high-resolution images but neglect the problem of domain shift. In contrast, we explicitly address domain shift by incorporating an energy-based loss derived from the negative energy score \citep{szekely2013energy, shen2024engression}. Second, while we adopt the energy-based loss from \citep{szekely2013energy, shen2024engression} to enhance the model extrapolability, our main contribution is extending this approach to multi-covariate settings and rigorously establishing theoretical consistency for estimation. Moreover, while domain shift has been studied in various contexts \citep{stacke2020measuring, zhang2021adaptive, liu2023few}, we are the first to address it in the context of multivariate outcomes. Although many domain adaptation regression methods have been proposed \citep{chen2021representation, wu2022distribution, taghiyarrenani2023multi}, they typically focus on univariate outcomes.


The remainder of this paper is organized as follows. Section~\ref{sec:related_works} presents related work on super-resolution for general and medical images. Section~\ref{sec:model} elaborates on the proposed model and theoretical results. Section~\ref{sec:learning_framework} illustrates the learning framework for 4DF data. Section~\ref{sec:4df_SR} presents the performance evaluation of the proposed model, and Section~\ref{sec:discussion} provides discussion of the results. Technical proofs and numerical simulations are provided in the Supplementary Material.

\section{Related Works}\label{sec:related_works}

{\bf General image super-resolution} aims to reconstruct a high resolution  image from its low resolution  version. Convolutional neural network (CNN)-based deep learning methods, such as SRCNN \citep{dong2014learning,dong2015image} and ESPCNN \citep{shi2016real}, learn a direct mapping from low resolution to high resolution images.  Furthermore,  SRResNet \citep{ledig2017photo} inherits residual learning from ResNet \citep{he2016deep}, while SRDenseNet \citep{tong2017image} adapts the dense blocks of DenseNet \citep{huang2017densely} for image super-resolution. Recently, attention mechanisms have been incorporated into super-resolution tasks, as demonstrated by models such as SwinIR \citep{liang2021swinir} and DRCT \citep{hsu2024drct}. Despite differences in architecture, these methods share a common regression-based framework and are typically trained using standard $L_1$ or $L_2$ loss functions.

\textbf{Medical image super-resolution} applies super-resolution techniques to medical imaging modalities such as magnetic resonance imaging (MRI) and computed tomography (CT). \cite{wang2016accelerating} introduced a CNN-based framework for MRI reconstruction, aiming to enhance image quality and reduce scanning time. Several 2D CNN-based approaches have since been proposed to improve MRI resolution using strategies such as progressive upsampling \citep{xue2019progressive}, multi-scale feature fusion \citep{qiu2021multiple}, and dual-branch architectures \citep{chen2022double}. More recently, generative adversarial network (GAN)-based models like SRGAN \citep{zhu2019can} and HiNet \citep{li2021high} have been developed to enhance MRI resolution for specific organs. In addition, \cite{wang2022adjacent} proposed a CNN that leverages adjacent slices for anisotropic 3D brain MRI super-resolution, which was further extended by \cite{guven2023brain} to dynamic contrast-enhanced volumes. \cite{georgescu2023multimodal} recently introduced a multimodal, multi-head convolutional attention module for super-resolving both MRI and CT data. However, these medical image super-resolution methods are primarily designed for structured volumetric data and are not directly applicable to irregular 3D geometries.

\textbf{4D Flow data super-resolution} aims to enhance high resolution velocity fields from low resolution 4DF data. \cite{ferdian20204dflownet} introduce a ResNet-based architecture, 4DFlowNet, to improve 4DF resolution. The model is trained on synthetic data generated from three aortic CFD simulations. This model has been later updated by \cite{ferdian2023cerebrovascular} using simulated cerebrovascular data from eight patient-specific geometries. Additionally, \cite{long2023super} proposed a physics-informed neural network trained on a single synthetic cardiovascular case to enhance 4DF resolution.  \cite{rutkowski2021enhancement} developed a super-resolution model trained on CFD data from five patient-specific cerebral aneurysm geometries. Unlike conventional image super-resolution benchmarks, these 4D Flow super-resolution approaches are typically trained on very limited datasets or simulated datasets, which poses significant challenges for out-of-domain generalization.

\section{Model and Theoretic Results}\label{sec:model}

\subsection{Multi-covariate pre-additive noise DSR model}\label{sec:eng_method}

Denote the high resolution CFD patch by $\b Y $ and let $\b X$ represent the low resolution data. For simplicity, we consider vectorized outcomes and covariates in this section. In practice, however, the shapes of the outcomes and covariates may vary.  We model the relationship between the high and low resolution data using a multi-covariate pre-additive noise DSR model, which assumes the following form:
\begin{align}\label{eq:eng_model1}
    \Y = \h\left( \X + \beps \right),
\end{align}
where $\h$ denotes an unknown mapping function and $\beps$ represents a Gaussian random vector that independent of $\b X$.  Intuitively, adding noise to the original domain artificially expands the input space, allowing the resulting estimator to naturally extend to a broader domain. More rigorously,  we demonstrate below that when the $\h$ takes specific semiparametric form that $\h(\t) \equiv \g(\bb\trans \t)$, employing a pre-additive noise model generally leads to better extrapolation performance compared to the standard regression model $\Y = \g(\bb\trans\X) + \beps$.  This result extends the findings of \cite{shen2024engression}, which considered the special case where $\bb = 1$ and a single covariate is considered.  Although the form $\g(\bb^\top \t)$ is more restrictive than the full nonparametric specification as it imposes a linear structure through $\bb^\top \t$, this framework can be readily extended to more flexible models where the inputs are constructed through basis expansion techniques such as splines \citep{de1978} and reproducing kernels \citep{wahba1990}, or are derived from the final layer of a deep learning network.

\subsubsection{Extrapolability of the DSR model}\label{sec:eng_extrapolability}

Consider an univariate outcome $Y$, Let $P(y\mid \b x)$ be the conditional distribution of $Y \mid \X= \b x$. Consider a class $\mathcal{P}=\left\{ P(y \mid \b x) \right\}$ of the conditional distributions. Define a distribution extrapolability measure as follows.
\begin{definition}[Distributional extrapolability]\label{def:distributional_extrapolability}
For $\delta > 0$, the distributional extrapolation uncertainly of model $\mathcal{P}$ is defined as
\begin{align*}
    \mathcal{U}_{\mathcal{P}}(\delta) := \sup_{{\b x}':d\left(\b x,{\b x}' \right) \le \delta} \sup_{ P, {P}' \in \mathcal{P}: \atop D_{\b x}(P,{P}')=0, \forall \b x \in \mathcal{X}} D_{{\b x}'}(P,{P}'),
\end{align*}
where $D_{\b x}(P,{P}'):=D\left\{ P(y \mid \b x), P'(y \mid \b x) \right\}$ for all $\b x \in \R^d$ and $D(P,{P}')$ denotes the distance between two probability distributions.  The class $\mathcal{P}$ is globally distributionally extrapolable if $\mathcal{U}_{\mathcal{P}}(\delta) \equiv 0$. Intuitively, a distribution class is extrapolable if, when two distributions in the class are close over a given covariate domain, they remain close when the domain is extended to a nearby closed neighborhood.
\end{definition}


The  model  class of interest is 
\begin{align*}
    \mathcal{M}= \left\{ g\left( \bm{\beta}^\top (\mathbf{x} + \bm{\epsilon}) \right) \mid g \in \mathcal{G}, \bm{\beta} \in \mathbb{R}^d, \beta_1 = 1 \right\},
\end{align*}
where 
\begin{align*}
    \mathcal{G} = \left\{ g : \mathbb{R} \to \mathbb{R} \mid g \text{ is monotone increasing with a finite derivative} \right\}
\end{align*}
and  $\bm{\beta}, \mathbf{x}, \bm{\epsilon} \in \mathbb{R}^d$. 
This formulation leads to the following theorem, which establishes distributional extrapolability of the model considering multi-covariates.

\begin{theorem}[Distributional extrapolability of DSR]\label{thm1}
Suppose that $g\left\{ \bm{\beta}^\top (\mathbf{x} + \bm{\epsilon}) \right\} \stackrel{d}{=} g'\left\{ \bm{\beta}'^\top (\mathbf{x} + \bm{\epsilon}) \right\}$ for all $\mathbf{x} \in \mathcal{X} \subset \mathbb{R}^d$, where $\bm{\beta}, \bm{\beta}' \in \mathbb{R}^d$ and $g, g' \in \mathcal{M}$. Then, the extrapolation uncertainly of model $\mathcal{M}$ satisfies $\mathcal{U}_{\mathcal{M}}(\delta) = 0$.
\end{theorem}

Theorem \ref{thm1} establishes that the model class $\mathcal{M}$ is extrapolable and the DSR model is in class $\mathcal{M}$. This suggests that the DSR model is robust to domain shift, and therefore can handle cases where covariates in the test data lie outside the range observed in the training data. 

\subsubsection{Finite-sample estimation}\label{sec:eng_estimation}

Consider an independent identifcally distributed sample $\left\{ (\X_i, Y_i) \right\}_{i=1}^n$ used during the training process. For each $i \in [n]$, we generate $m$ independent samples of the noise variable $\bm{\epsilon}$, denoted as $\left\{ \bm{\epsilon}_{i,j} \right\}_{i,j=1}^{n,m}$. The mapping function $g$ is estimated by
\begin{align}\label{eq:est_g}
    \hat{g}, \wh{\bb} \in \mathrm{argmin}_{g \in \mathcal{M}} {\mathcal{L}}(g, \bb),
\end{align}
where
\be\label{eq:lossuni}
   {\mathcal{L}}(g, \bb) &\equiv& \frac{\sum_{i=1}^n \sum_{j=1}^m \left| Y_i - g \left\{ \bm{\beta}^\top (\X_i + \bm{\epsilon}_{i,j}) \right\} \right|}{nm} \nonumber\\
    && - \frac{1}{2 nm(m-1)}\sum_{i=1}^n \sum_{j=1}^m \sum_{j'=1}^m \left| g \left\{ \bm{\beta}^\top (\X_i + \bm{\epsilon}_{i,j}) \right\}-g \left\{ \bm{\beta}^\top (\X_i + \bm{\epsilon}_{i,j'}) \right\} \right|.
\ee
We show that minimizing (\ref{eq:lossuni}) is equivalent to minimizing the Cramér distance, which is defined as the $L_2$ distance between the estimated and empirical cumulative distribution functions, as stated in Proposition 1 at Appendix. This equivalence forms the basis for our distributional learning method. The following theorem demonstrates the consistency of the finite-sample estimation under specified conditions.  

\begin{theorem}[Estimation Consistency]\label{thm2}
Let $\mathcal{X} \equiv \{\x = (x_1, \ldots, x_d)\trans, x_j \in [-k_j, k_j], \|\x\|_2<\infty\}$ and let $F_{Y \mid X; g, \bm{\beta}}(y, \mathbf{x})$ denote the true cumulative distribution function of $Y = g\left( \bm{\beta}^{\top} \X + \bm{\beta}^{ \top} \bm{\epsilon}_{i,j} \right)$ given $\X$, where $\left\{ \bm{\epsilon}_{i,j} \right\}_{i,j=1}^{n,m}$ are i.i.d. random errors, $\g$ and $\bb$ are true mapping function and parameter vector, respectively. Assume that the density corresponding to $F_{Y \mid X; g, \bm{\beta}}(y, \mathbf{x})$ is bounded away from zero and that $|Y_i|$ is bounded above for all $i= 1, \ldots, n$. Let $\wh{g}$ and $\wh{\bm{\beta}}$ denote the estimated function and parameter vector, respectively, obtained via \eqref{eq:est_g}. Furthermore, assume the following conditions hold.
\begin{enumerate}[label=(C\arabic*)]
  \item \label{con:c0}  The noisy $\bm{\epsilon}$ follows a distribution that is symmetric around zero with a bounded support.

  \item \label{con:c1}  Let $Q_{\alpha}^{\b z}(\bb)$ denote the $\alpha$-th quantile of random variable $\bb^\top \b Z$. Assume $\{x_j \b e_j, j = 1, \ldots, d, x_j \in [-k_j, k_j ]  \} \in \mathcal{X}$. And there is a $\b x_j \in \{x_j\b e_j, x_j \in [-k_j, k_j]  \}$  such as $\bm{\beta}^{\top} \b x_j + Q_\alpha^{\bm{\epsilon}}(\bm{\beta}) \in \{\bm{\beta}^{\top} \b x_j^*, \b x_j^* \in \mathcal{X}_j\}$ for all $\alpha \in [0, 1]$. 

  \item \label{con:c2}  $\mathcal{G}$ is a function set containing bi-Lipschitz and strictly monotone functions with finite derivative. 
  
  \item \label{con:c3}  The observed convariates lies on grid points $\mathcal{X}_n = (\b x_1, \ldots, \b x_K)$. Furthermore, $n_k = \sum_{i=1}^n I(\X_i =
  \b x_k) \to \infty$ and $K/n \to 0$. Let $n_{\min} = \min(n_k, k = 1, \ldots, K)$. Assume $ K = O\{\log(n_{\min})\}$ and $n_{\min} \to \infty$. Furthermore, assume that $\max_{k_1, k_2, k_1\neq k_2}\|\b x_{k_1} - \b x_{k_2}\|_2 = O(K^{-1})$. And for any $\b x \in \mathcal{X}$, there is an $x_k \in \mathcal{X}_n$ such that $\|\b x -\b x_k\|_2\leq d_K = O(K^{-1})$. 
  
  \item \label{con:c4}  For any $\bm{\beta}$, we can choose $\alpha_s \in [0, 1]$ for $s = 1, \ldots, p$, $p>d$, such that 
  \begin{align*}
        \b M (\bm{\beta}) = \left[ \frac{\partial
        Q_{\alpha_s}^{\bm{\epsilon}}(\bm{\beta})}{\partial \bm{\beta}}  - \left(\sum_{j=1}^d \beta_j\right)^{-1} {\bm{1}}, s = 1, \ldots, p  \right] ,
  \end{align*}
  satisfies $0 <c \leq \left\| \b M(\bm{\beta}) \right\|_2/\sqrt{p} \leq C $.
\end{enumerate}
Then we have
\begin{align*}
    \left\| \wh{\bm{\beta}} - \bm{\beta} \right\|_2 & \overset{p}{\rightarrow} 0, & \left| \hat{g}(z) - g(z) \right| \overset{p}{\rightarrow} 0,
\end{align*}
for all $z \in \mathbb{R}$ as $n \to \infty$.
\end{theorem}

\begin{remark}
    The symmetry assumption in Condition~\ref{con:c0} ensures that the median quantile satisfies $Q_{0.5}^{\bm{\epsilon}}(\bb) = 0$ for any vector $\bb$. Moreover, the bounded support condition imposed on the noise $\bm{\epsilon}$ guarantees the feasibility of Condition~\ref{con:c1}, as it ensures that the random variable $\bm{\beta}^{\top} \b x_j + Q_\alpha^{\bm{\epsilon}}(\bm{\beta})$ lie within the set $\{\bm{\beta}^{\top} \b x_j^*, \b x_j^* \in \mathcal{X}_j\}$. Condition~\ref{con:c2} is a standard assumption as adopted in \citep[Condition (B2)]{shen2024engression}. By leveraging the bi-Lipschitz continuity condition of $\mathcal{G}$, we establish consistency outside the observed domain, which is generally unattainable with traditional regression approaches. Condition~\ref{con:c3} is imposed primarily for analytical convenience; we believe that same results would remain valid if $\b X$ were assumed to follow a general continuous density. Condition~\ref{con:c4} is also a mild assumption according to the boundedness of $\beps$.
\end{remark}

The technical proofs of our theoretical results are provided in Appendix. It is worth noting that the composite function $g \circ \bb\trans$ can be approximated by many deep learning architectures, such as the 3D U-Net used in our paper. In this design, the first convolutional layer is linear and corresponds to the operator $\bb\trans$, while the nonlinear function $g$ is approximated by the subsequent layers of the network.


\subsubsection{Multivariate outcome with a deep learning framework}\label{sec:multi}

When considering a multivariate outcome, we extend the loss function in (\ref{eq:lossuni}) to the more general form in (\ref{eq:lossmulti}), where we use a deep learning framework  to approximate $\h$. Specifically, we write the parameterized model as $\b Y = \h_{\btheta}(\b X + \beps)$, where $\btheta$ includes the parameters of the deep learning architecture used to approximate $\g$ and the parameter $\bb$.  The estimator for $\btheta$, denoted by $\wh{\btheta}$, is then obtained by minimizing ${\mathcal{L}}(\btheta)$ in (\ref{eq:lossmulti}). Using this estimator, our upsampling function is given by $E\left\{\h_{\wh{\btheta}} \left( \X_i + \beps_{ij} \right)\right\}$, where the expectation is taken over $\beps_{ij}$ and can be appoximated by the empircal expectation. 
\be\label{eq:lossmulti}
    {\mathcal{L}}(\btheta) &\equiv &\frac{\sum_{i=1}^n \sum_{j=1}^m \left\| \Y_i - \h_{\btheta} \left( \X_i + \beps_{ij} \right) \right\|_2}{nm} \nonumber\\
    && - \frac{1}{2 nm(m-1)}\sum_{i=1}^n \sum_{j=1}^m \sum_{j'=1}^m \left\| \h_{\btheta} \left( \X_i + \beps_{ij} \right)-\h_{\btheta} \left( \X_i + \beps_{ij'} \right) \right\|_2.
\ee

\section{Distributional Super-Resolution Famework}\label{sec:learning_framework}

In this section, we describe our proposed distributional super-resolution framework tailored for 4DF data. Section~\ref{sec:data_preparing} outlines the data preparation procedure prior to model training. Section~\ref{sec:down_sample} introduces our self-supervised pre-training approach. Section~\ref{sec:model_fine_tune} details the subsequent fine-tuning methodology, and Section~\ref{sec:model_pred} describes the model prediction step.

\subsection{Data cropping, augumentation and downsampling}\label{sec:data_preparing}

To ensure DSR is invariant to blood vessel geometry, we propose training and testing the model on local patches instead of the entire vessel structure. Let $\s_j := (s_{j,x}, s_{j,y}, s_{j,z}) \in \mathbb{R}^3$ be the voxel location for the $j$ voxel and $\v_j := (v_{j,x}, v_{j,y}, v_{j,z}) \in \mathbb{R}^3$ be the velocities in three directions recorded at each voxel $\s_j$. The following steps describe how to construct the velocity cubic patch from velocity data on a 3D vessel structure.
\begin{itemize}
    \item \textbf{Step 1. Neighborhood identification}: Define the $\epsilon$-ball as 
    \begin{align*}
        \mathcal{N}_{\epsilon} (\s_{j}) = \left\{  \s_{k}: \left\| \s_{k} - \s_{j} \right\|_2 \le \epsilon, k =1,\ldots, N \right\},
    \end{align*}
    where $\epsilon >0$ is a tunable parameter representing the neighborhood radius, $N$ is the  number of independent voxels. For each spatial location $\s_j$, the adaptive resolution parameter is calculated via
    \begin{align*}
        t_j := \left \lfloor \frac{\log_2 \left( \left| \mathcal{N}_{\epsilon} (\s_j) \right| \right)}{3} \right \rfloor,
    \end{align*}
    where $\left| \cdot \right|$ denoting cardinality
    and $\left \lfloor \cdot \right \rfloor $ denote the floor function.  
    
    \item \textbf{Step 2. Voxel grid construction}: 
    Let $\mathcal{B} = \left[ x_{\min},x_{\max} \right] \times \left[ y_{\min},y_{\max} \right] \times \left[ z_{\min},z_{\max} \right]$ be the minimal axis-aligned bounding box containing $\mathcal{N}_{\epsilon} (\s_j)$. 
    The bounding box $\mathcal{B}$ is partitioned into a $2^{t_j} \times 2^{t_j} \times 2^{t_j} $ regular grid, resulting in $2^{t_j} \times 2^{t_j} \times 2^{t_j} $ voxels. We denote the tensor containing the centroids of the cubes formed by the grid points as $\b G^j = \left[ G_{mnk} \right]_{m,n,k \in [t_j]} \in \R^{2^{t_j} \times 2^{t_j} \times 2^{t_j} \times 3}$ and $G_{mnk} \in \R^3$.

    \item \textbf{Step 3. Velocity assignment}: Let $\b V^j = \left[ V_{mnk} \right]_{m,n,k \in [t_j]} \in \R^{2^{t_j} \times 2^{t_j} \times 2^{t_j} \times 3}$ denote the velocity tensor associated with centroid grid $\b G^j$. Each element $V_{mnk} \in \R^3$ is determined through adaptive spatial averaging:
    \begin{align}\label{eq:construct_velocity_in_voxel}
        V_{mnk} = \begin{cases}
             \v_i & \text{if }  \exists i, \ \text{s.t.} \ \left\| \s_i -G_{mnk} \right\|_2 = 0 \ \text{for} \ \s_i \in \mathcal{N}_{\epsilon} (\s_j) \\
             \sum_{i=1}^{I} w_i \v_i  & \text{otherwise} 
            \end{cases}
    \end{align}
    where $I = \left| \mathcal{N}_{\epsilon} (\s_j) \right|$ and the weights $w_i$ satisfy:  
    \begin{align*}
        w_i = \frac{\left\| \s_i - G_{mnk} \right\|_2^{-1}}{\sum_{l=1}^I \left\| \s_l - G_{mnk} \right\|_2^{-1}}, \quad \forall \s_i, \s_l \in \mathcal{N}_{\epsilon} (\s_j),
    \end{align*}
    and $\v_i$ is velocity asscociated to $\s_i$. 
    In summary, when a centroid $G_{mnk}$ coincides with any $\epsilon$-ball memeber $\s_i$, that point's velocity $\v_i$ inherits the voxel velocity directly; for non-coincident voxels, velocities are computed as a linear combinations of neighborhood velocities, with weights inversely proportional to Euclidean distances.

    \item \textbf{Step 4. Tensor resizing}: Because the deep learning architecture requires inputs of equal dimension, we set  $T:=\max_{j \in [N]} t_j$ and extend each velocity tensor $\b V^j \in \R^{2^{t_j} \times 2^{t_j} \times 2^{t_j} \times 3}$ to $\b V^j \in \R^{2^T \times 2^T \times 2^T \times 3}$ via nearest-neighbor imputation.
\end{itemize}
After constructing the velocity tensor $\b V^j$ using Steps 1-4, we generate the input–output training pairs by applying a downsampling procedure to each $\b V^j$. To further augment the dataset, we repeat downsampling $L$ times. And then  we construct input–output pairs at every downsampling level.  The complete procedure is described in the supplement material.

\begin{figure}
    \scriptsize
		\begin{center}
			\begin{tabular}{c}			\includegraphics[width=1.0\textwidth]{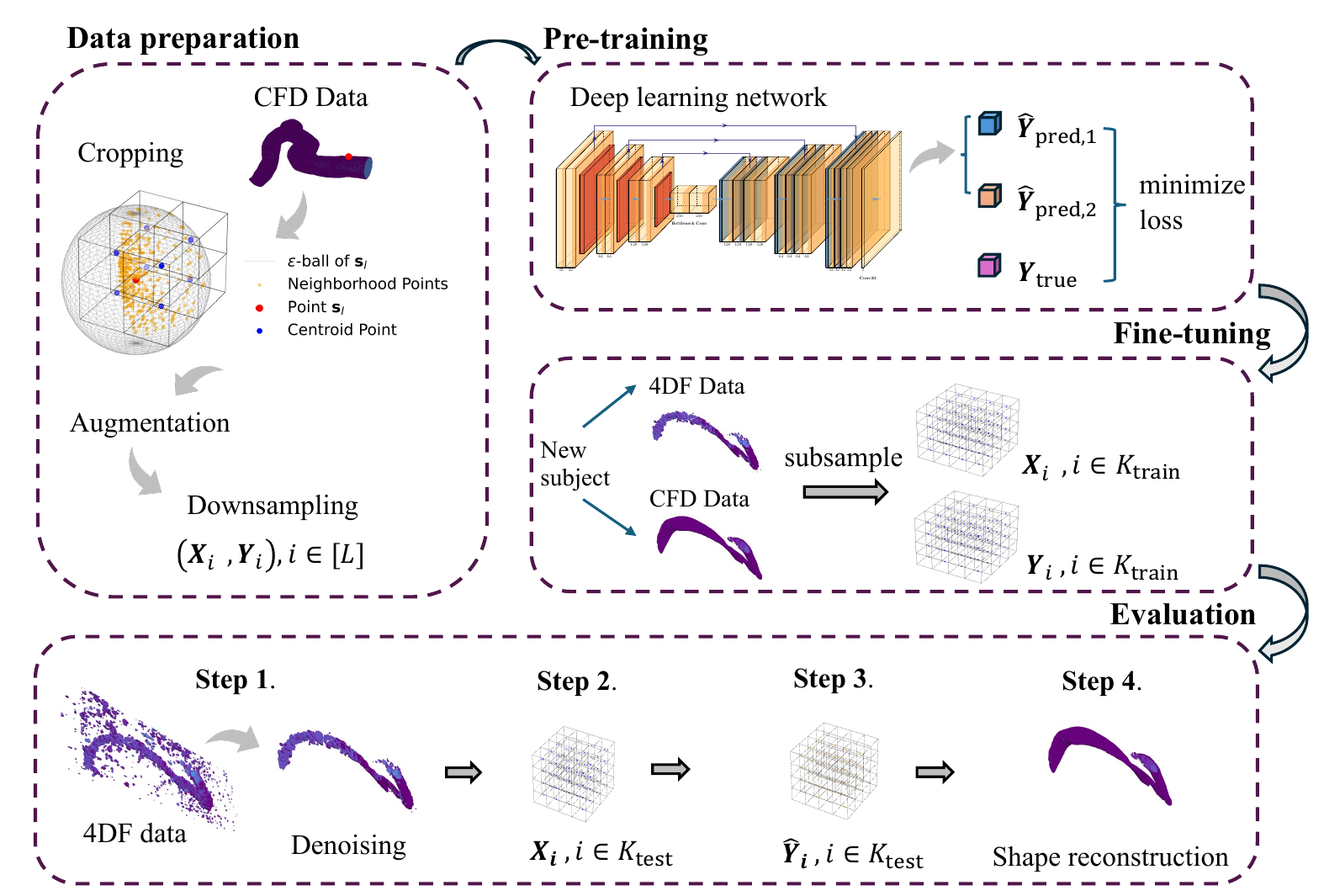}
				\\
			\end{tabular}
		\end{center}
\caption{Flow chart for distributional super-resolution procedure. Data Preparation: CFD training pairs are generated through cropping, data augmentation and downsampling.
Pre-training: These pairs are used to pre-train the proposed DSR model.
Fine-tuning: A small subset of paired 4DF and CFD patches is used to refine the model's estimators.
Evaluation: The fine-tuned model is then evaluated on the complete 4DF testing dataset. }
    \label{fig:flowchart_process}
\end{figure}

\subsection{Self-supervised learning to establish a pre-train model}\label{sec:down_sample}

\graphicspath{{figs1/}}
\begin{figure}
    \begin{center}
        \begin{tabular}{c}	
        \psfig{figure=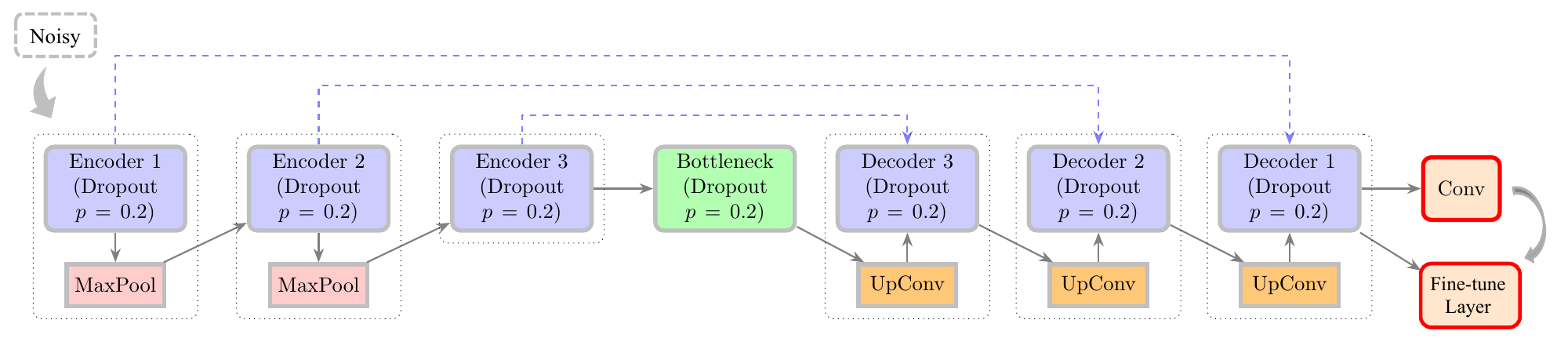,width=1.0\textwidth,angle=0} \\
        \end{tabular}
	\end{center}
    
    \caption{The 3D U-Net architecture. The convolutional layer in the red boxes is replaced by new layer in the fine-tuning process. }
    \label{fig:3d_unet_structure}
\end{figure}
We employ a 3D U-Net architecture \citep{cciccek20163d} in Figure \ref{fig:3d_unet_structure} to represent $\h_{\btheta}$, and estimate the network parameters $\btheta$ by minimizing the loss function defined in (\ref{eq:lossmulti}). Specifically, for each input tensor $\X_i$, we add a randomly sampled perturbation vector $\beps_{ij}$ drawn from a centered normal distribution to create a noise-perturbed input. The 3D U-Net is then applied to this perturbed input to produce one output. This process is repeated $m$ times for each input, generating $m$ outputs, which are used to construct the loss function in (\ref{eq:lossmulti}). Finally, we train the network by minimizing the loss with the Adam optimizer, using a learning rate of $10^{-4}$ to obtain the pre-trained estimator of the network parameters.

\subsection{Model fine-tune}\label{sec:model_fine_tune}

In the fine-tuning phase, we follow the two-step strategy proposed by \cite{kumar2022fine}, which combines linear probing with full fine-tuning (LP-FT).  Specifically, the final convolutional layer of the pre-trained 3D U-Net was replaced with a designed fine-tune layer. Initially, all layers except the fine-tune layer were frozen, and the fine-tune layer alone was trained for 300 epochs. Subsequently, all layers were unfrozen, and the entire network further training for an additional 200 epochs. An illustration of the two-step fine-tuning strategy is provided in Figure~\ref{fig:3d_unet_structure}.

The dataset used for fine-tuning consists of patch-wise CFD and 4DF pairs. These pairs are extracted from full fluid measurements over a 3D intracranial blood vessel geometry. The extraction of cubic patches follows Steps 1–2 outlined in Section~\ref{sec:data_preparing}, with voxel locations defined based on the 3D structure used for CFD simulations. Due to noise in the 4DF data, velocity measurements outside the blood vessel can be small but are not necessarily zero. To mitigate the impact of this noise, we use the 3D structure from the CFD simulations to identify the smallest cubic region that contains all valid geometric points on the blood vessel, and discard the remaining noisy measurements outside this region \textbf{Step 1 in Figure~\ref{fig:flowchart_process}}. Note that the 3D geometrical structure can be readily constructed from contrast-enhanced MRA images \citep{liu2021identification}.

Following Steps~1–2 in Figure~\ref{fig:flowchart_process}, we partition the intracranial blood vessels into $K = 99$ cubic patches, each containing velocity data from both CFD and 4DF. Of these, 15 patches are used for fine-tuning the model, while the remaining patches serve as the test set for performance evaluation.

\subsection{Prediction}\label{sec:model_pred}

In this section, we describe the model prediction procedure on the test data. The process involves two main steps: (1) performing patch-wise predictions to upsample the local resolution of the 4DF data, and (2) using the resulting densely upsampled patches to interpolate velocity values onto a predefined 3D blood vessel structure.

\begin{itemize}

    \item \textbf{Upsampling on grid}: 
    For each test sample, the high resolution representation is computed as
    \begin{align*}
        \hat{\b Y}_i = \frac{1}{J} \sum_{j=1}^J \left\{ \h_{\wh{\btheta}} \left(\b X_i + \bm{\epsilon}_{ij} \right) \right\},
    \end{align*}
    where $\bm{\epsilon}_{ij} \sim \mathcal{N}(\bm{0}, \sigma^2 \b I)$ represents independent noise, $J$ is the predefined number of samples and $\h_{\wh{\btheta}}$ is the estimated function resulting from the fine-tuning process. Note that the value of 
    $J$ does not need to match the samples of noise used during training. To balance the computational cost and prediction accuracy, we set $J = 100$ in our estimation process. The variance term $\sigma^2$ is selected to optimize the prediction performance in the fine-tuning procedure. The estimator, denoted as $\wh{\btheta}$, is obtained by minimizing (\ref{eq:lossmulti}).

    \item \textbf{Shape reconstruction}: 
    The final prediction results are provided on regular 3D grids, which may not fully encompass all spatial locations on the 3D blood vessel geometry. To address this, we implement a shape reconstruction (\textbf{Step 4} in Figure~\ref{fig:flowchart_process}), for each location on the predefined 3D geometry, we interpolate the velocity values using a nearest-neighbor approach based on the densely upsampled data from \textbf{Step 4} in Figure~\ref{fig:flowchart_process}.
    
\end{itemize}

\section{Simulation Study}\label{sec:sim}

In this section, we present two simulation studies to examine the extrapolation property of the proposed DSR model. These experiments focus on evaluating extrapolation property.  Section~\ref{sec:sim_low_dim} and Section~\ref{sec:sim_high_dim} present simulations with synthetic data under low and high dimensional covariate settings, respectively. 
Across both simulations, the proposed DSR model shows strong extrapolatability and consistently outperforms the standard regression model on test data whose domain deviates from the training sample.

\subsection{Low dimensional  setting}\label{sec:sim_low_dim}

In this simulation, we consider the input $\left\{\b x_{i}, i = 1, \ldots, n \right\}$, where each $\b x_i$ is a three-dimensional vector generated from $n$ independent samples. For $i$, we generate $\x_i$ as
\begin{align*}
    \b x_i &= \left( x_{\mathrm{upper}} - x_{\mathrm{lower}} \right) \b u_i + x_{\mathrm{lower}} \in \R^3, & \b u_i \sim \mathcal{U}(0,1)^3 \in \R^3,
\end{align*}
where $\mathcal{U}(a,b)$ denotes the uniform distribution on the interval $(a,b)$, and $x_{\mathrm{upper}}$ and $x_{\mathrm{lower}}$ represent the upper and lower bounds of the training support, respectively. We set the true coefficient vector as $\bm{\beta} = (1,1.2,1.5)^\top$.
We then generate independent zero-mean noise vector $\beps_i $ with the covariance-variance matrix $\b \Sigma = \diag \left\{1 /\var(\bb\trans \x_i) \right\} \in \R^{3 \times 3}$. 
Finally, we generate the univariate target $y_i \in \R$ as
\begin{align*}
    y_i = g \left\{ \bm{\beta}^\top \left(\b x_i + \bm{\epsilon}_i \right) \right\}. 
\end{align*} In this paper, we assess the performance of DSR under four different mapping functions $g$ defined as  follows
\begin{itemize}
  \item[] \begin{tabular}{@{}l l}
    \textbf{softplus}: $g(x) = \ln (1+e^x)$; & 
    \textbf{square}: $g(x) = \{\max(0,x)\}^2/c_1$;
  \end{tabular}
  \item[] \begin{tabular}{@{}l l}
    \textbf{log}: $g(x) = \begin{cases}
      \tfrac{x}{3} + \ln (3) - \tfrac{2}{3}, & x \le 2 \\
      \ln(1+x), & x > 2
    \end{cases}$; &
    \textbf{cubic}: $g(x) = x^3/c_2$.
  \end{tabular}
\end{itemize}
where $c_1 = 7.4$ and $c_2 = 11.1$. Here, $c_1$ and $c_2$ are scaling factors chosen to normalize the functions to a comparable scale.
During the training process, we generate $n = 1000$ samples for each simulation, with the input support bounded between $x_{\mathrm{lower}} = 0$ and $x_{\mathrm{upper}} = 1.08$. 
In the training data, we utilize the neural network model to estimate $g\circ\bb\trans$. 

We evaluate the accuracies of the DSR estimators on new inputs $\b x_{\mathrm{eval},i} \in \R^3$ that potentially outside the training domain. To simplify the evaluation, we assume that all dimensions of $\b x_{\mathrm{eval},i}$ take the same value and define  
\begin{align}\label{eq:xeval}
    x_{\mathrm{eval},i,j} = (i-1)\frac{1.64}{n-1}, \quad i = 1, \ldots, n, j = 1, 2, 3. 
\end{align}
Since the upper bound of $x_{\mathrm{eval}, i, j}$ is 1.64, which exceeds $x_{\mathrm{upper}}$, some $x_{\mathrm{eval}, i, j}$ values lie outside the training domain.  The prediction using the true parameters is  calculated by
\begin{align}
    y_{\text{pred},i} = \frac{1}{N} \sum_{m=1}^N g \left\{ \bb\trans\left( \b x_{\mathrm{eval},i} + \beps_m \right) \right\}\label{eq:true_eval_mean}, 
\end{align}
and the prediction using the estimated parameters is 
\begin{align}
    \wh{y}_{i} = \frac{1}{N} \sum_{m=1}^N \wh{g} \left\{ \wh{\bb}\trans\left( \b x_{\mathrm{eval},i} + \beps_m \right) \right\}\label{eq:predicted_eval_mean}, 
\end{align}
where $\wh{g}$ and $\wh{\bb}$ are estimated via deep learning using the training data, $N=10000$ and $\left\{ \beps_m \right\}_{m \in [N]}$ are generated following the same procedure as in the training process.

We also estimate $g$ and $\bb$ using a standard regression model of the form $Y = g(\bb^\top \X) + \beps$. The predictions are then given by $\wt{y}_i = \wt{g}(\wt{\bb}^\top \x_{\rm{eval}, i})$, where $\wt{g}$ and $\wt{\bb}$ are learned with a neural network model under the mean squared error loss. We refer to this approach as $L_2$ regression. 

\begin{figure}
    \begin{center}
       \begin{tabular}{cc}				\psfig{figure=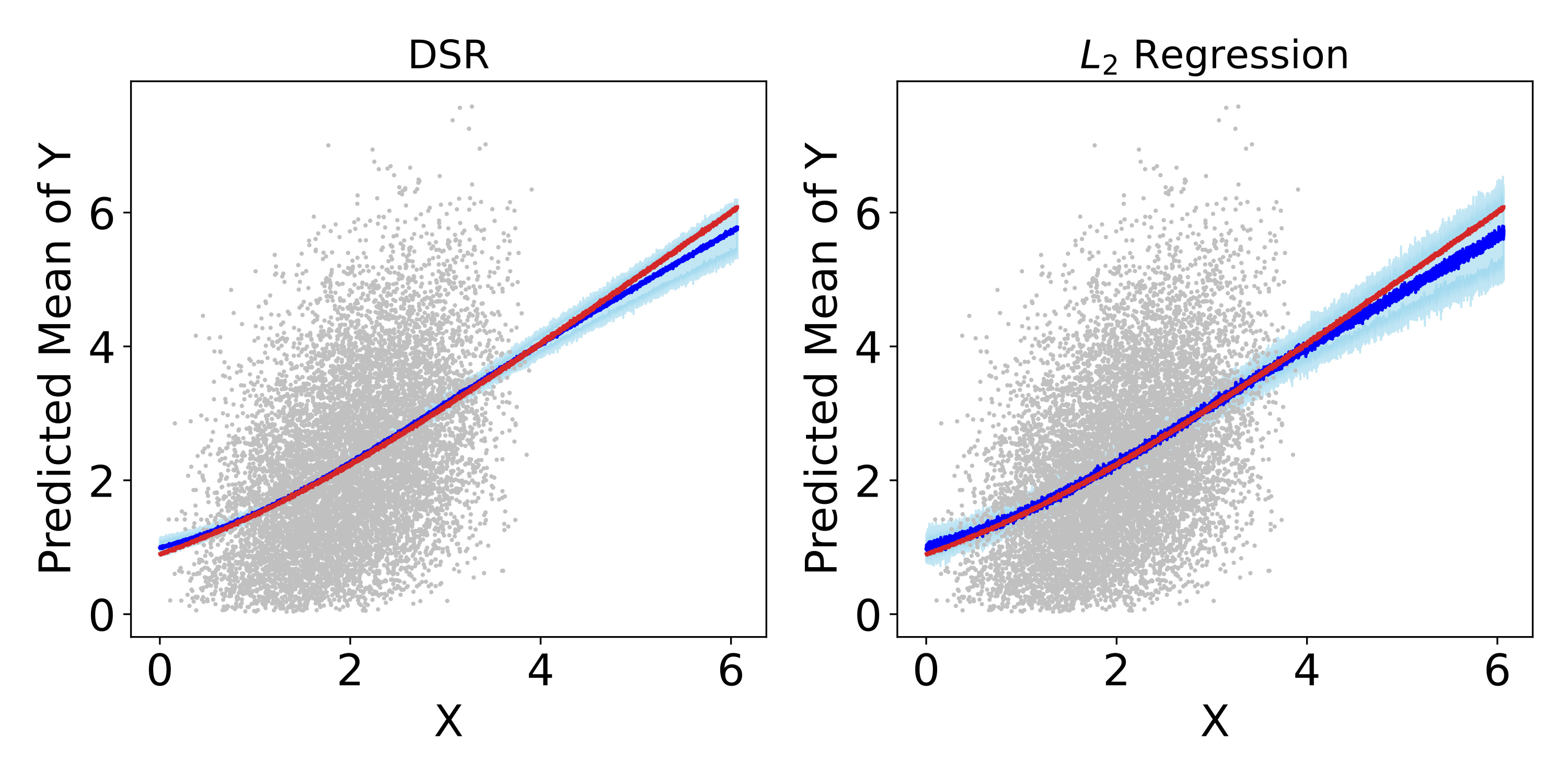,width=0.4\textwidth,angle=0} & \psfig{figure=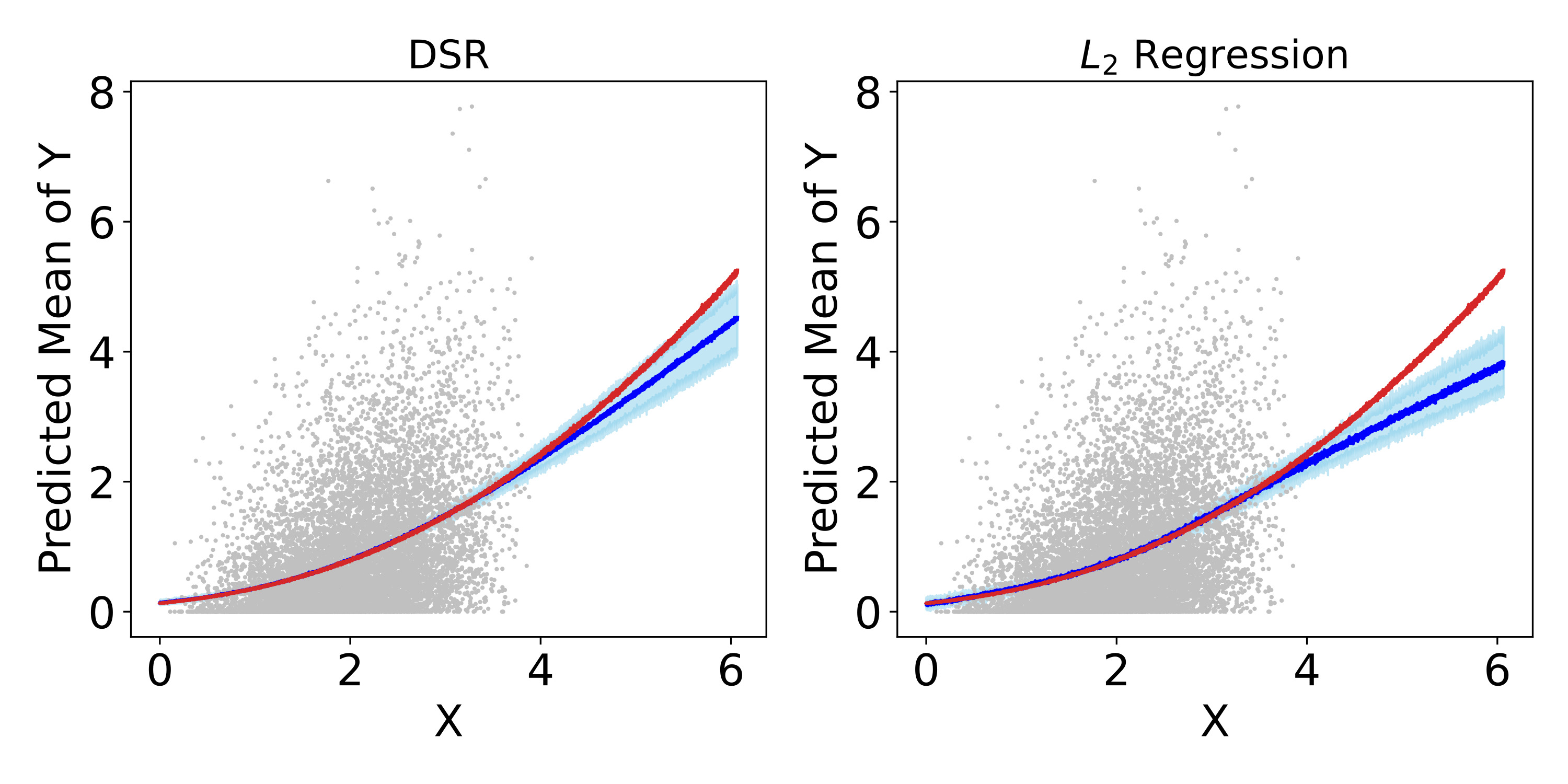,width=0.4\textwidth,angle=0} \\
       (A) Softplus & (B) Square \\
        \psfig{figure=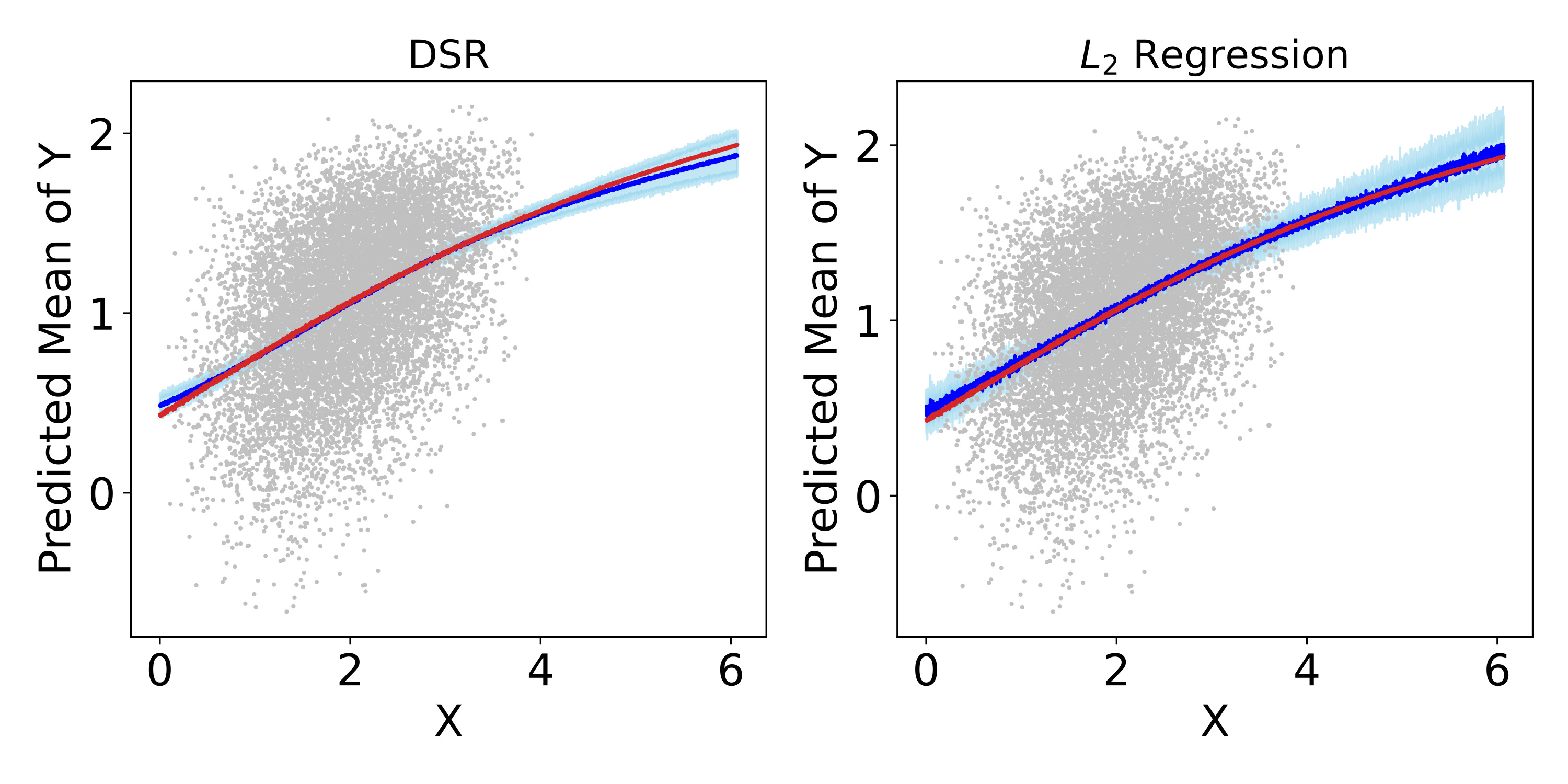,width=0.4\textwidth,angle=0} & \psfig{figure=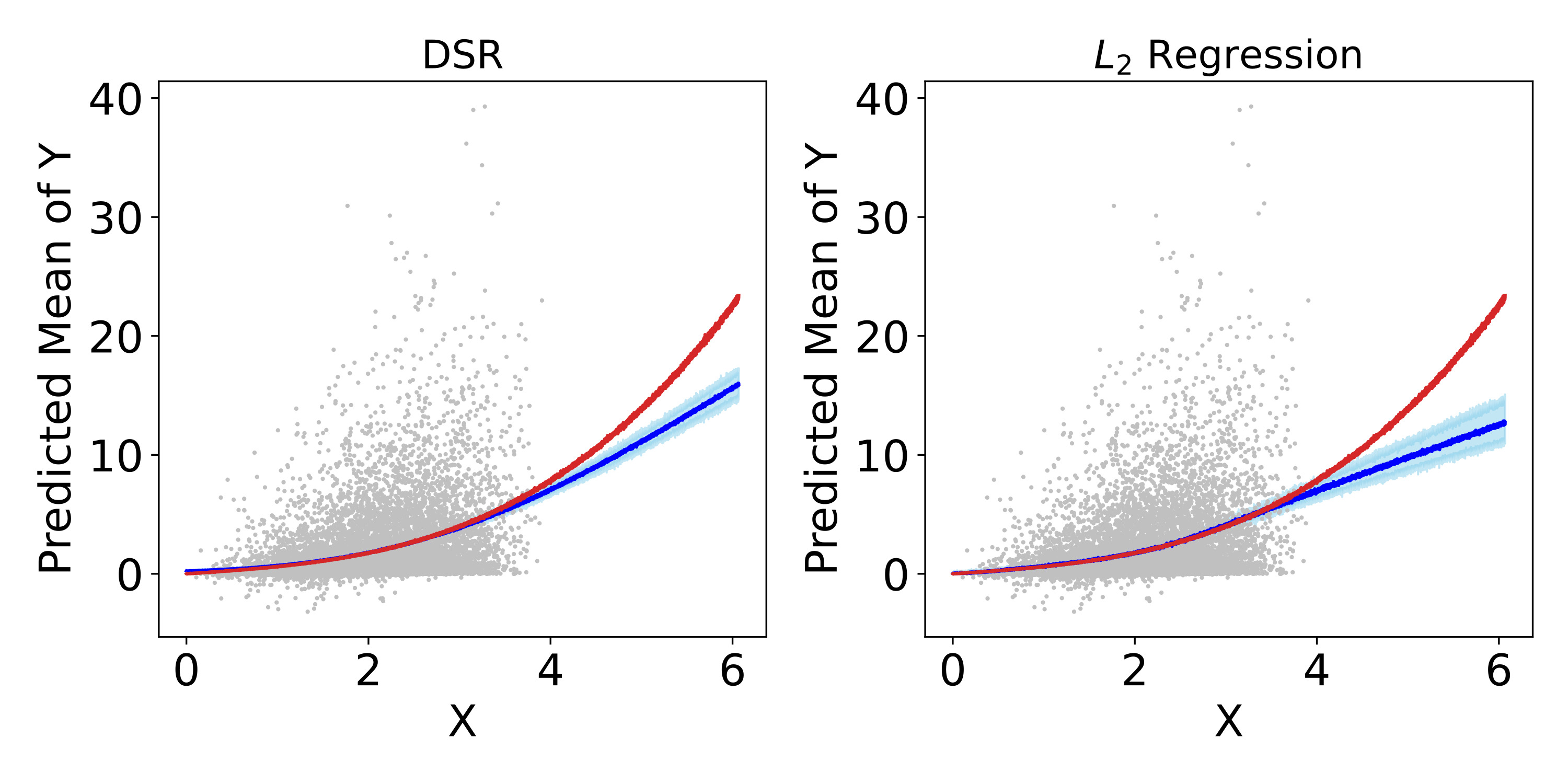,width=0.4\textwidth,angle=0} \\   

        (C) Log & (B) Cubic \\
        \end{tabular}
	\end{center}
    
    \caption{The comparision between DSR and $L_2$ regression models under the low dimensional setting. The gray dots denote the training data, red line denotes the prediction with true parameters, blue line denotes the prediction with estimated parameters from DSR or $L_2$ regression models and the shaded areas denote the 10\% to 90\% quantile interval.}
    \label{fig:compare_sim1}
\end{figure}

We conduct 20 simulation runs, and the results are presented in Figure~\ref{fig:compare_sim1}. In each plot, the red line represents $y_{\mathrm{pred}, i}$ versus $\bb^\top \b x_{\mathrm{eval}, i}$, while the blue line shows the average of $\wh{y}_{i}$ or $\wt{y}_{i}$ across the 20 simulations versus $\bb^\top \b x_{\mathrm{eval}, i}$. The shaded regions are the confidence bands, which indicate the 10\% and 90\% quantiles of $\wh{y}_{i}$ or $\wt{y}_{i}$ versus $\bb^\top \b x_{\mathrm{eval}, i}$. For reference, we also include scatter plots of $y_i$ against $\bb^\top \b x_i$ from the training data, illustrating that the training domain is narrower than the testing domain.  As illustrated in Figure~\ref{fig:compare_sim1}, for the square and cubic functions, the mean predictions produced by the DSR model align more closely with the true mean than those of the $L_2$ regression model, particularly in regions where the testing data fall outside the training domain. Furthermore, the DSR model also outperforms $L_2$ regression model under softplus and logarithmic functions with narrower confidence bands.

\graphicspath{{figs1/}}

\subsection{High dimensional setting}\label{sec:sim_high_dim}

\begin{figure}
    \begin{center}
        \begin{tabular}{cc}				\psfig{figure=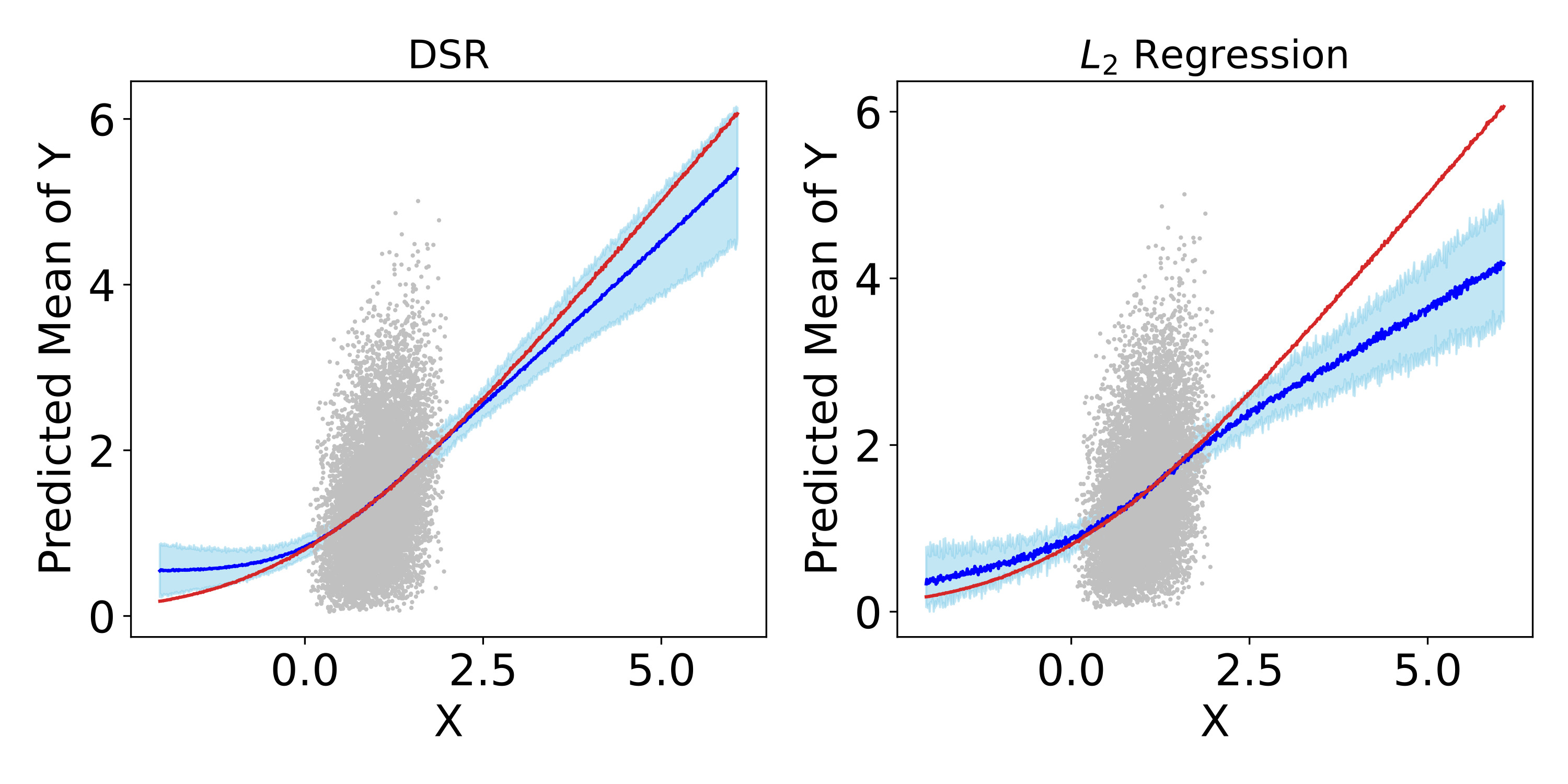,width=0.4\textwidth,angle=0} & \psfig{figure=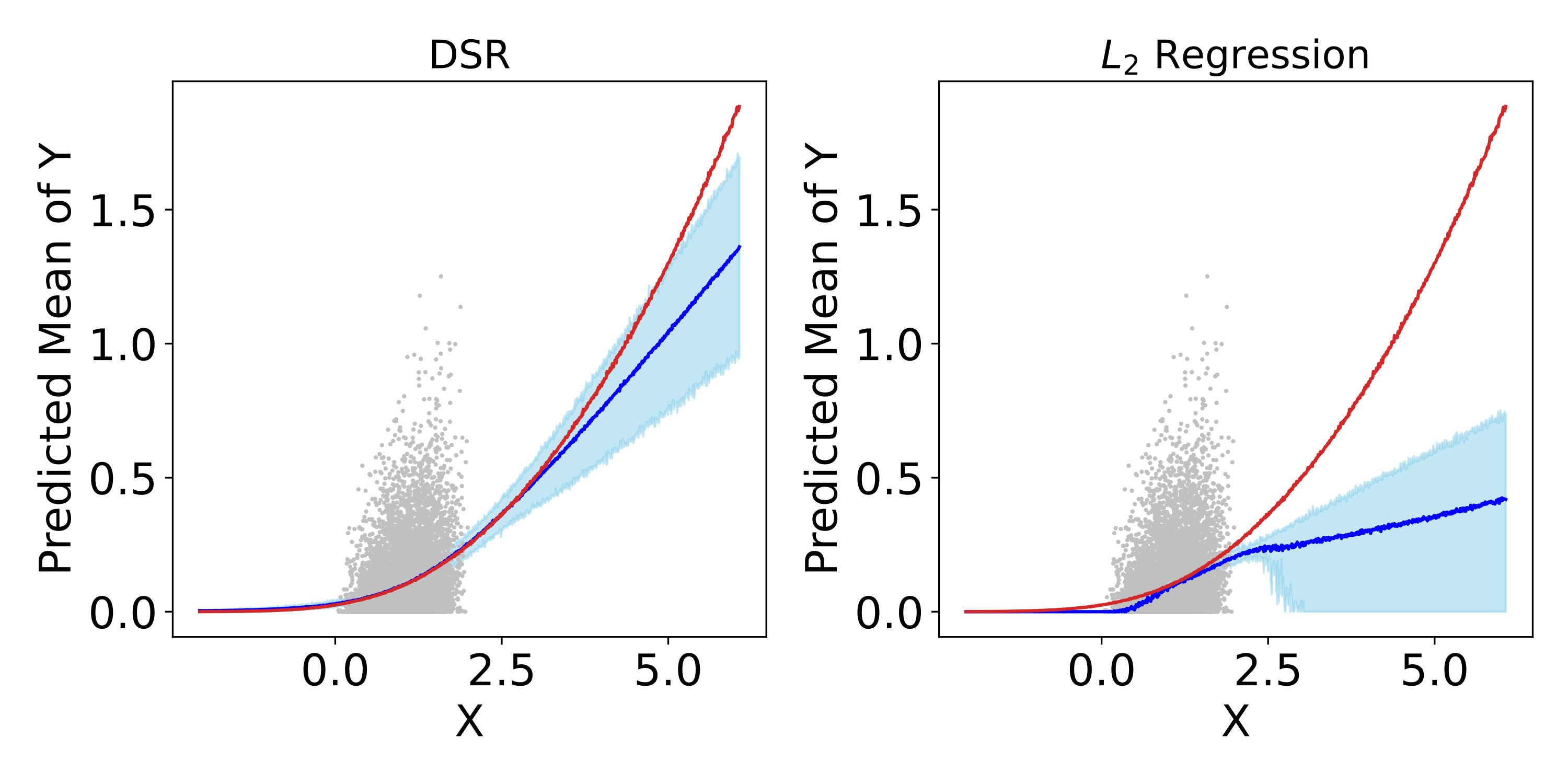,width=0.4\textwidth,angle=0} \\
        (A) Softplus & (B) Square \\
        \psfig{figure=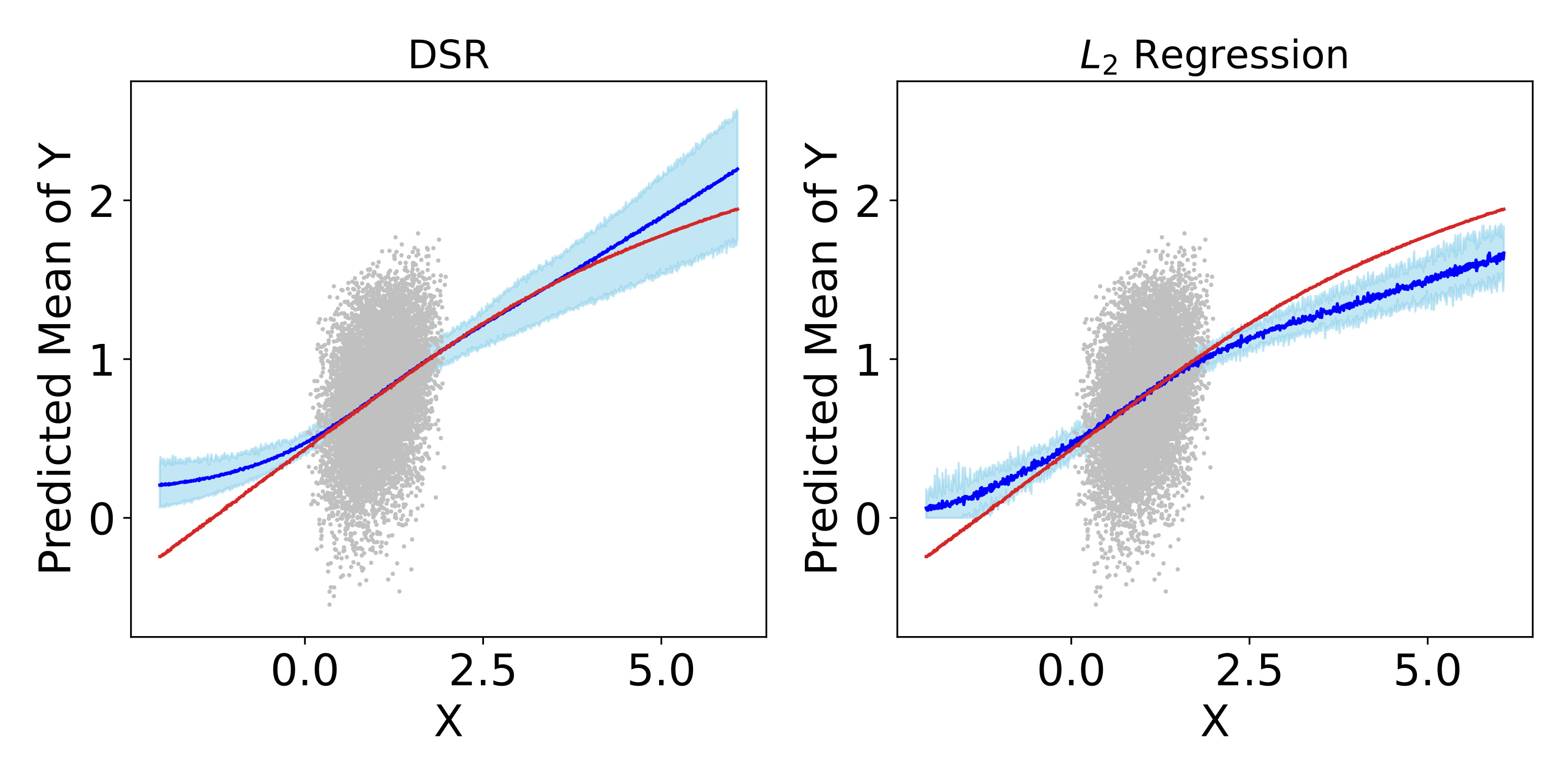,width=0.4\textwidth,angle=0} & \psfig{figure=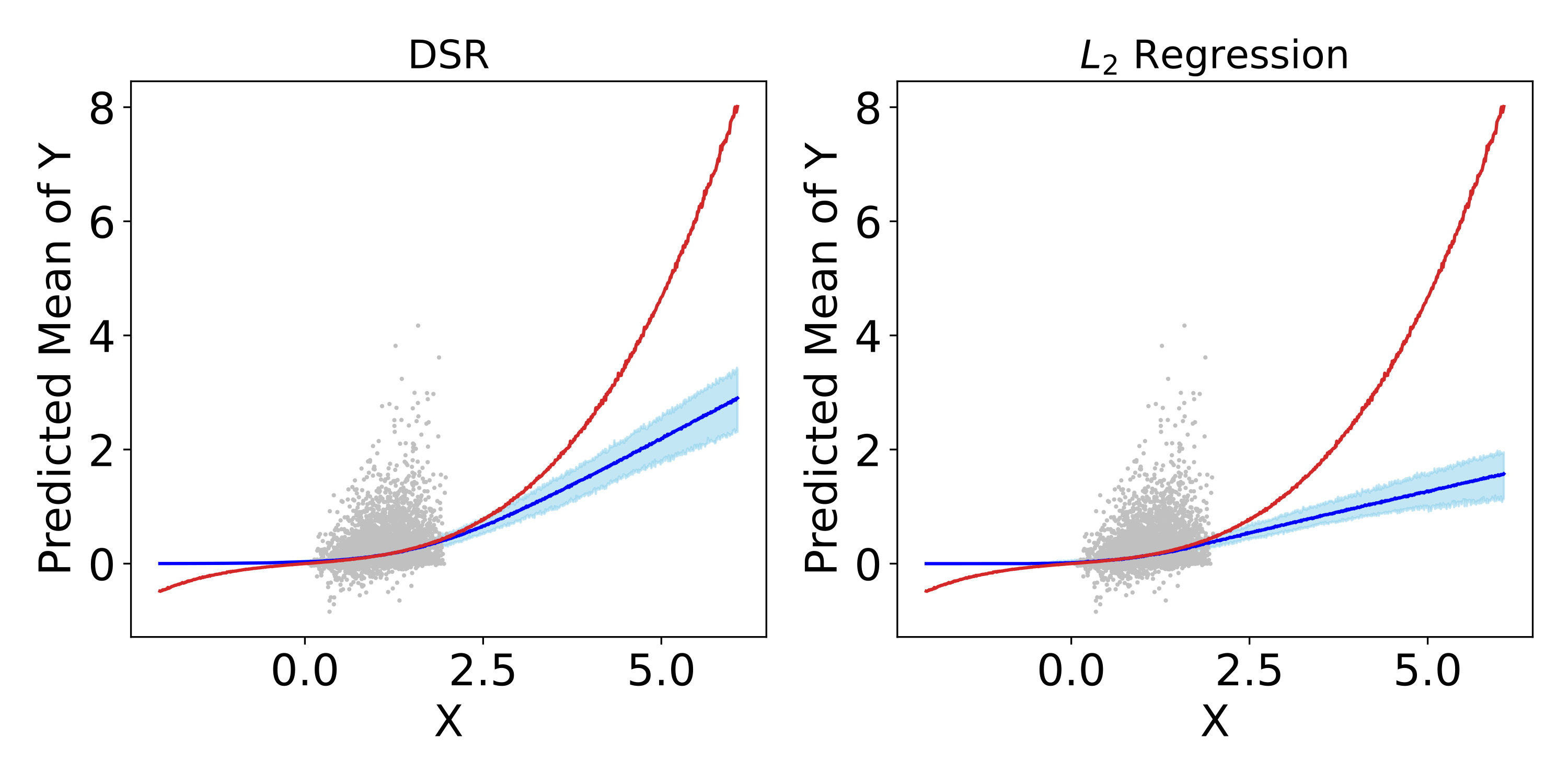,width=0.4\textwidth,angle=0} \\   

       (C) Log & (B) Cubic \\
    \end{tabular}
	\end{center}
    
    \caption{The comparision between DSR and $L_2$ regression models under the high dimensional setting. The gray dots denote the training data, red line denotes the prediction with true parameters, blue line denotes the prediction with estimated parameters from DSR or $L_2$ regression models and the shaded areas denote the 10\% to 90\% quantile interval.}
    \label{fig:compare_sim2}
\end{figure}

In this section, we consider a high-dimensional covariate setting, where each input vector $\b x_{i}$ is 64-dimensional. We define a sparse coefficient vector $\bm{\beta} \in \R^{64}$, with the three randomly selected entries given by $(1,\,1.2,\,1.5)$ and all remaining entries set to zero. The input vectors $\x_i$ is generated as in the low-dimensional setting. The noise term $\beps_i$ are generated separately for the nonzero- and zero-coefficient subsets. For the nonzero subset, the noise $\beps_i \sim \mathcal{N}(0, \b \Sigma)$ where $\b \Sigma = \diag(1/\|\bb\|_2^2) \in \R^{3 \times 3}$. For the zero subset,  $\beps_i$ follows a standard multivariate normal distribution. 
The $g$ functions are chosen from softplus, square, log, and cubic, as before. For the square and cubic functions, we set $c_1 = 20$ and $c_2 = 30$, respectively.
We then follow the DSR framework to generate the true outcomes $y_i$. For evaluation, we construct the dataset $\b x_{\mathrm{eval},i} \in \R^{64}$, where all dimensions share the same value as defined in (\ref{eq:xeval}).
Figure \ref{fig:compare_sim2} presents a comparison between the DSR model and the $L_2$ regression model based on $y_{\rm{pred}, i}$, $\wh{y}_i$, and $\wt{y}$ as defined in Section~\ref{sec:sim_low_dim}. The results show that the DSR model consistently outperforms the $L_2$ regression model, yielding more accurate predictions across all four mapping functions.

\graphicspath{{figs1/}}

\section{Real data analysis: 4DF Super-Resolution}\label{sec:4df_SR}

Our pre-train dataset comprises \( n = 1200 \) independent 3D patches of CFD data. We split this dataset into training (1000) and validation subsets (200) to estimate the parameters of the DSR model. Additionally, we obtained 99 paired 3D patches of CFD and 4DF data. Of these, 15 pairs are used for fine-tuning the model, while the remaining pairs serve as the test set to evaluate DSR performance and compare it with benchmark methods.



\subsection{Compare the performance of DSR and  $L_2$ regression model on the pre-training data. }

\begin{figure}[!h]
    \centering
    \includegraphics[width=1.0\textwidth]{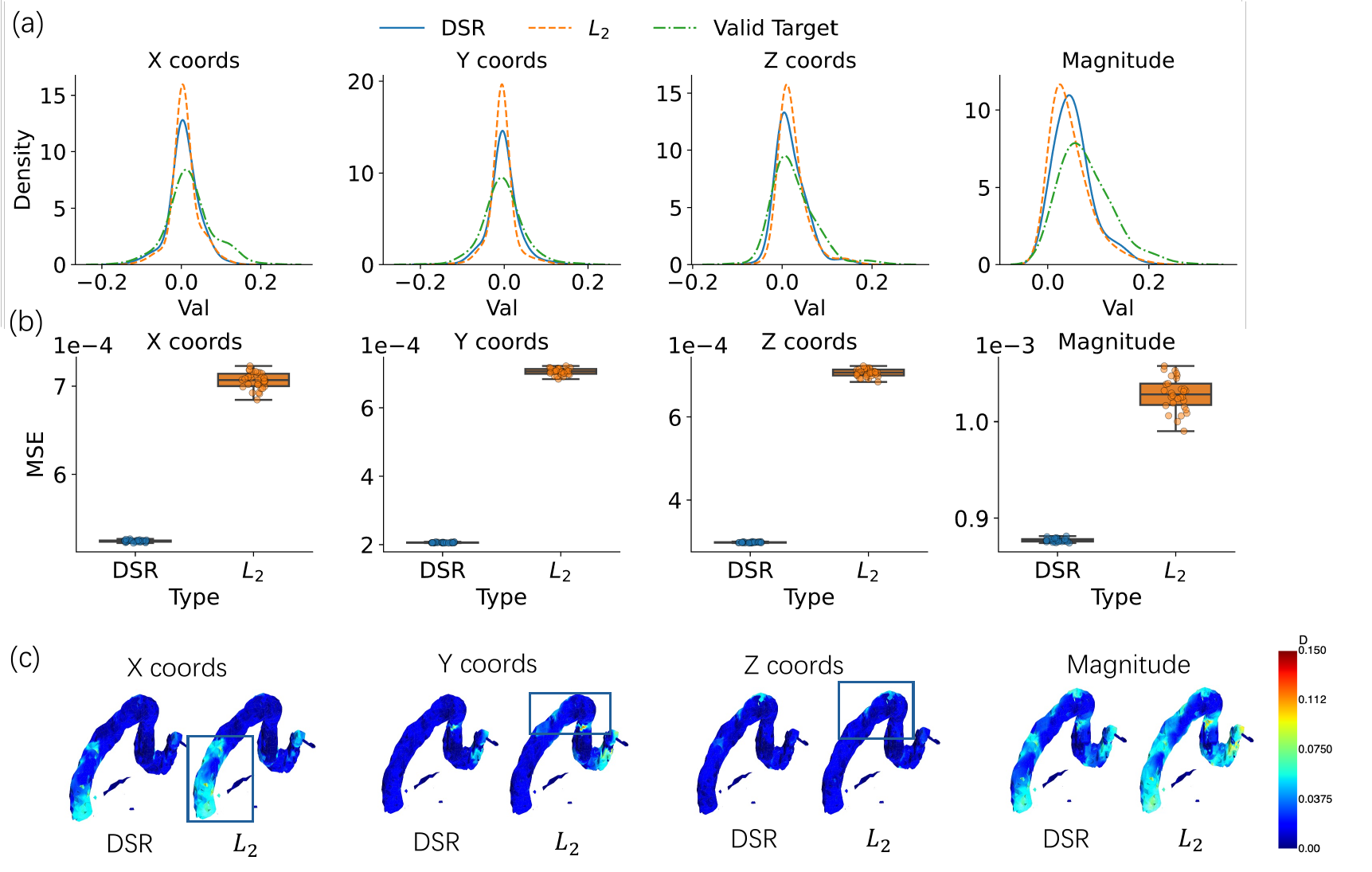}\\

    \caption{The performance of DSR and $L_2$ regression model on the validation data.   The magnitude is defined as the vector norm of the values along the three spatial directions. (a) Comparison of fitted density estimates between the DSR model and the $L_2$ model. (b) MSE comparison between the prediction error of DSR model and the $L_2$ model. (c) Absolute error heatmaps for a representative subject. Blue box regions in which  DSR achieves lower errors than the $L_2$ model baseline.} 
    \label{fig:compare_pretrained_valid}
\end{figure}

During the pre-training process, we train DSR for 200 epochs by minimizing (\ref{eq:lossmulti}). Furthermore, we compare the prediction results on the validation CFD data against those obtained using the $L_2$ regression model.
As shown in Figure~\ref{fig:compare_pretrained_valid}(a), on the validation data, the distribution of predictions from DSR is closer to the true values than that from the $L_2$ regression model. Furthermore, Figure~\ref{fig:compare_pretrained_valid}(b) illustrates that DSR achieves smaller prediction errors than the $L_2$ regression model in every direction, and in magnitude, which is   defined as the $L_2$ norm of the three component dimensions.

We also highlight regions on the 3D geometry in Figure~\ref{fig:compare_pretrained_valid}(c) where DSR outperforms the $L_2$ regression model. The results show that DSR achieves higher prediction accuracy, particularly at vessel bifurcations and inlet regions. It is worth noting that if the training and validation domains completely overlap, the DSR and $L_2$ regression models yield similar results. As shown in Figures~\ref{fig:compare_sim1} and \ref{fig:compare_sim2}, when the testing domain lies within the training domain, the predictions from the two models are nearly identical.
 The superior performance of DSR on the CFD validation data suggests that, even within the CFD pre-train dataset, there exists a certain level of domain shift. This is likely due to the fact that the training and varlidation data are drawn from different subjects.

\graphicspath{{figs1/}}

\subsection{Comparison with benchmarks after fine-tuning
}
We fine-tune DSR using samples consisting of paired high-resolution CFD data and low-resolution 4DF inputs. A set of 15 patches is used for fine-tuning, with each low-resolution 4DF input paired with its corresponding CFD target. Fine-tuning is performed for 500 epochs until convergence. The fine-tuned DSR is then tested on 84 patches from the testing dataset. 

We compare the fine-tuned DSR model with a model trained using the $L_2$ regression approach and with the 4DFlowNet model proposed by \citep{ferdian20204dflownet}, which demonstrated strong performance on synthetic 4DF data. All three models are pre-trained and fine-tuned on the same datasets. The primary difference between DSR and the benchmarks is that DSR accounts for potential domain discrepancies between the training and testing data by adding artificial noise to the input data to enlarge the training domain and by employing the distributional loss function described in (\ref{eq:lossmulti}). In contrast, the $L_2$ regression and 4DFlowNet models do not account for this possibility, instead relying on standard regression and using the mean squared error loss to estimate the parameters.

\begin{figure}
    \begin{center}
        \begin{tabular}{c}	
        \psfig{figure=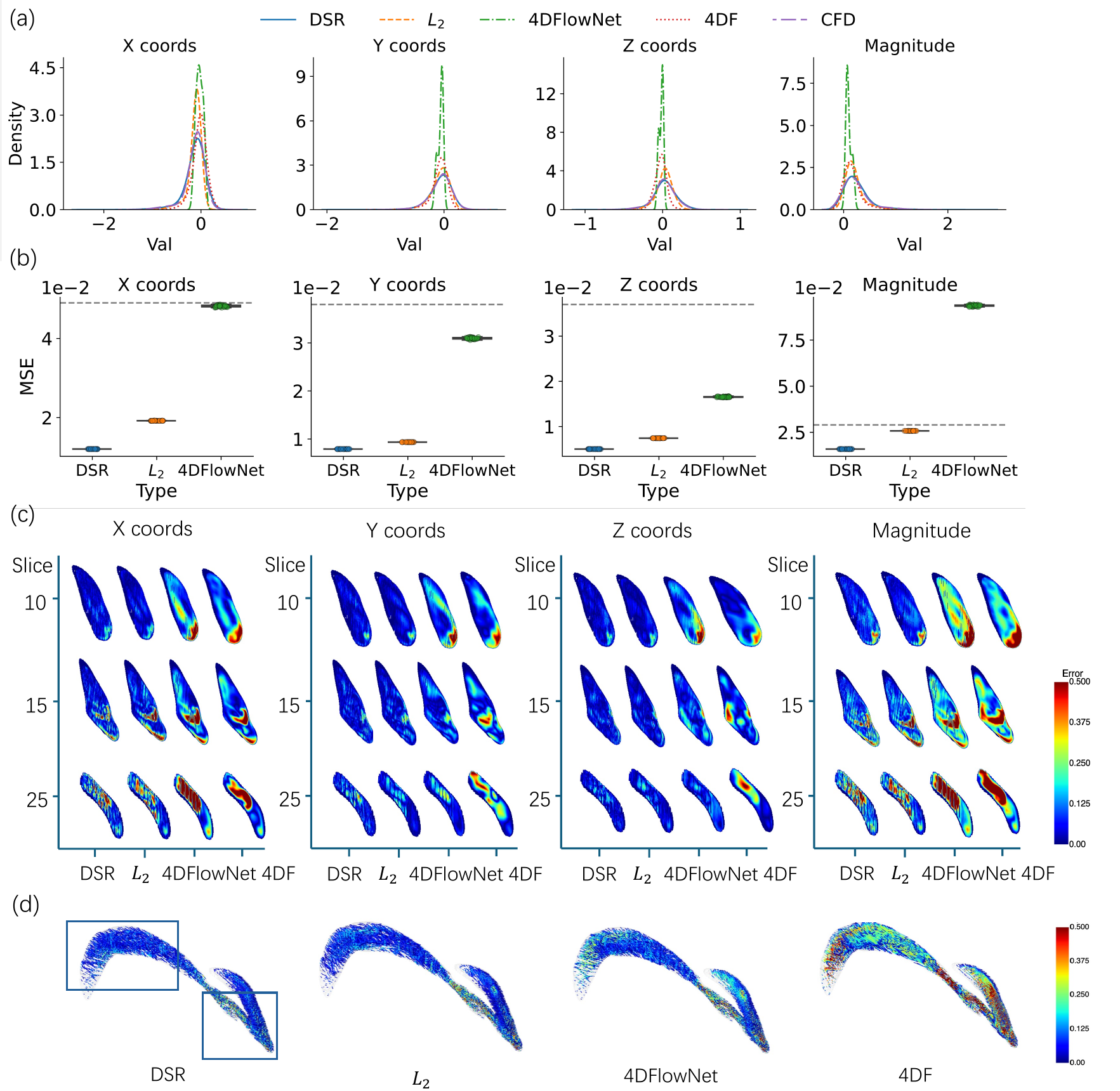,width=1.0\textwidth,angle=0} \\
        \end{tabular}
	\end{center}
    
    \caption{The comparison of DSR fine-tuned model, $L_2$ fine-tuned model and 4DFlowNet fine-tuned model. (a) Comparison of fitted density estimates between the fine-tuned models, high resolution CFD data and low resolution 4DF data. (b) MSE comparison between the prediction error of three fine-tuned models. The gray dotted lines denote the MSE between low resolution 4DF and high resolution CFD. (c) Heatmaps of the prediction error at slices $10, 15, 25$. (d) Estimation errors on vascular geometry.}
    \label{fig:compare_tune_model_exp1}
\end{figure}

As shown in Figure~\ref{fig:compare_tune_model_exp1}(a), the distributions of the predictions from the DSR and the $L_2$ regression models closely match the true distribution, whereas 4DFlowNet predictions exhibit noticeable deviations from the truth. Furthermore, Figure~\ref{fig:compare_tune_model_exp1}(b) shows that 4DFlowNet predictions have significantly larger errors compared to the predictions from the DSR and $L_2$ regression model. The DSR model, in particular, achieves substantially lower errors than the $L_2$ regression model, especially in the $X$ and $Y$ directions. Overall, DSR consistently attains the lowest mean squared errors across all dimensions relative to both the $L_2$ regression and 4DFlowNet models. Moreover, the MSE between the velocities from the low-resolution 4DF data and the high-resolution CFD data is substantially higher than the MSE between the outputs of any of the three models and the high-resolution CFD data. This underscores both the effectiveness and necessity of the super-resolution process.

In Figure~\ref{fig:compare_tune_model_exp1}(c), we visualize the prediction errors at slices 10, 15, and 25 for all three models, showing that DSR achieves the highest prediction accuracy on those slides. Finally, Figure~\ref{fig:compare_tune_model_exp1}(d) presents the estimation errors on the real vascular geometry. The results demonstrate that DSR outperforms the other methods, particularly in regions with high vessel curvature.

\graphicspath{{figs1/}}

\subsection{Impact of pre-training and data augmentation on model performance}

\begin{figure}
    \begin{center}
       \begin{tabular}{c}				    
        \psfig{figure=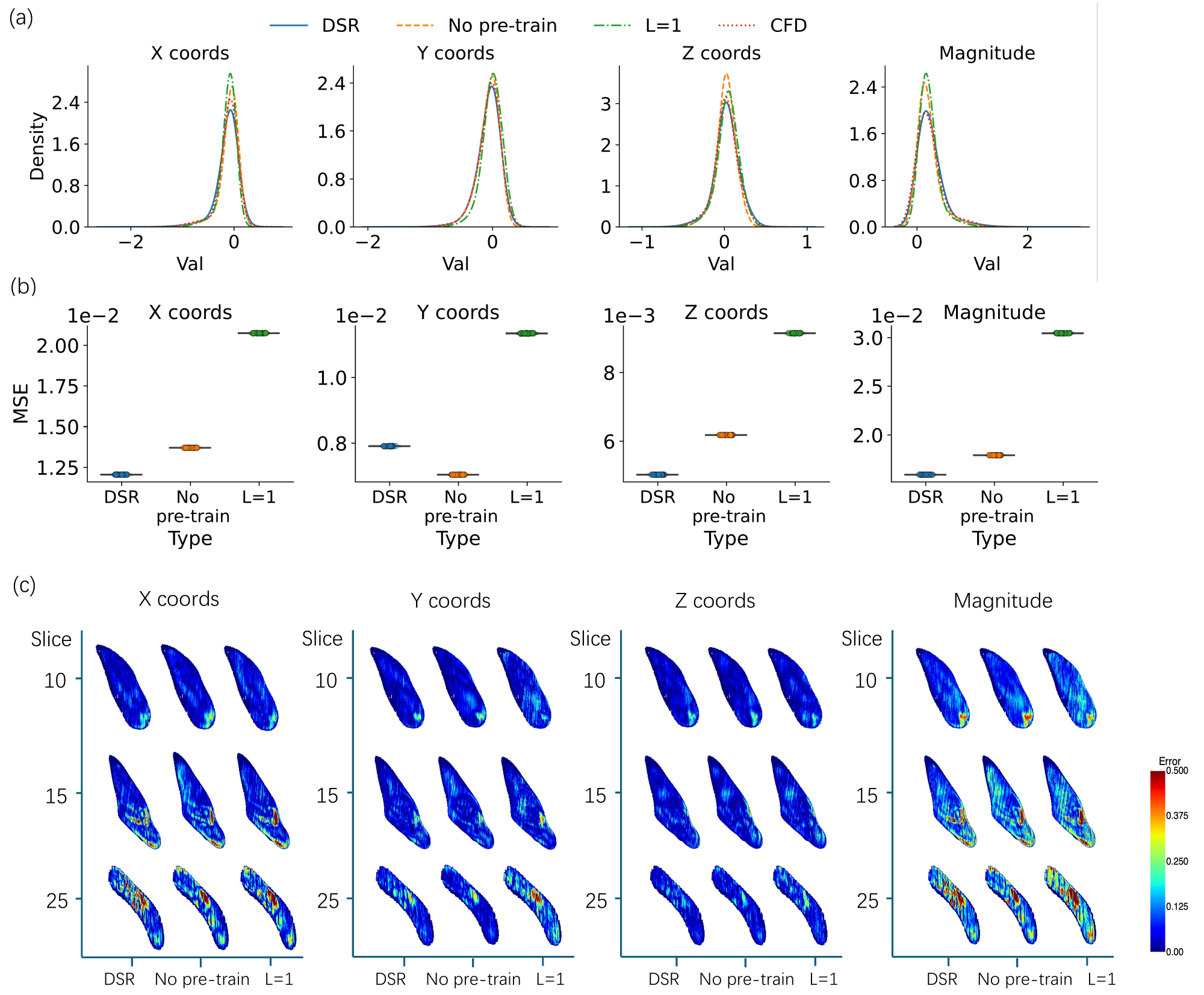,width=1.0\textwidth,angle=0} 
        \end{tabular}
	\end{center}
    
    \caption{Evaluation of pre-training and data augmentation effects in DSR model}
    \label{fig:compare_diff_DSR_model}
\end{figure}

To evaluate the benefit of pre-training, we also train DSR directly on the fine-tuning datasets without pre-training. This experiment examines whether pre-training provides a useful initialization that enhances downstream performance. The resulting prediction MSE is shown with the label ``No pre-train'' in Figure~\ref{fig:compare_diff_DSR_model}.
It is clear that the model with pre-training outperforms the one trained without it. This result underscores the importance of pre-training, especially when the available datasets are limited.

In the current DSR framework, we perform data augmentation by downsampling the CFD data four times ($L = 4$), thereby enlarging the pre-training dataset. To demonstrate the benefit of this augmentation, we compare it with a model trained using data downsampled only once ($L = 1$). Our results indicate that DSR trained with data from four-stage downsampling consistently outperforms the model trained with data from a single downsampling stage. As shown  in Figure~\ref{fig:compare_diff_DSR_model}, the model trained without augmentation ($L=1$) exhibits higher prediction MSE than the one trained with multi-stage augmentation ($L = 4$), highlighting the effectiveness of this strategy.

\graphicspath{{figs1/}}

\subsection{Robustness analysis of fine tuning procedure}

\begin{figure}
    \begin{center}
        \begin{tabular}{c}				    
        \psfig{figure=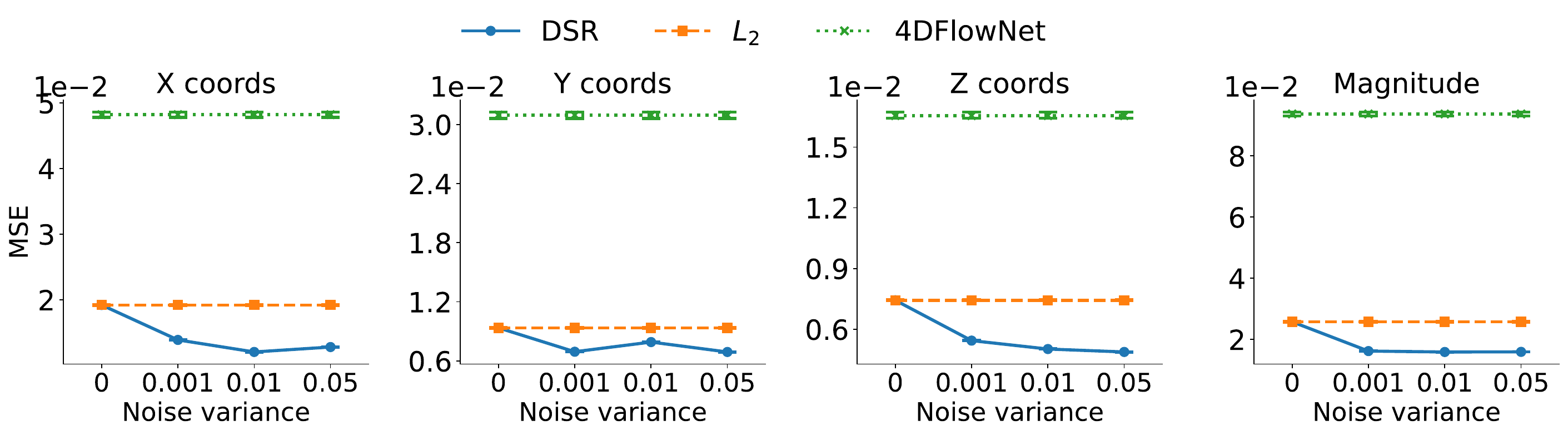,width=1.0\textwidth,angle=0} \\
        (a) Comparison of the models across different noise levels.  \\
        \psfig{figure=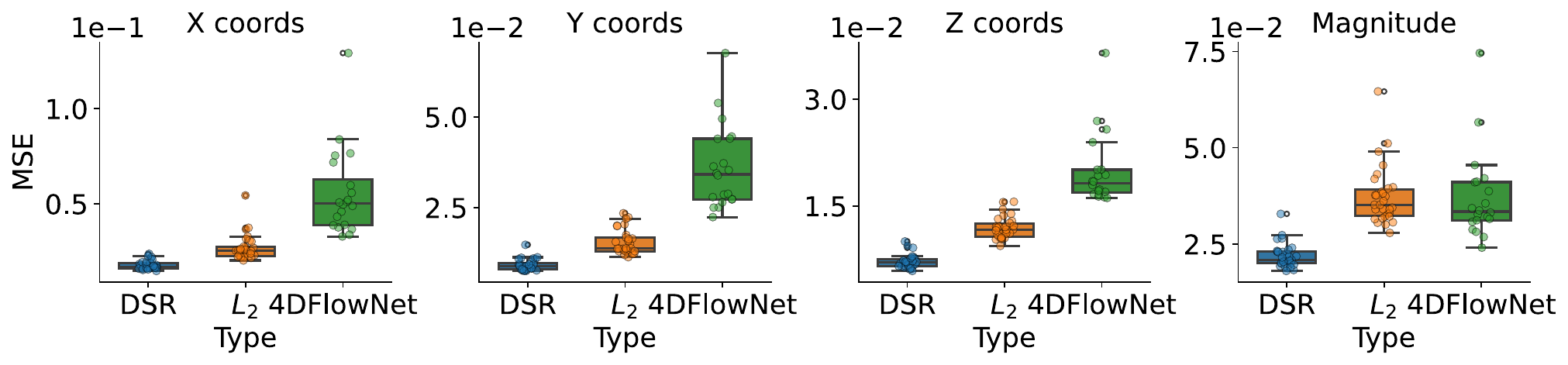,width=1.0\textwidth,angle=0} \\
        (b) Comparison of methods across different resampling sets.
        \end{tabular}
	\end{center}
    
    \caption{Robustness analysis of the fine-tuning process}
    \label{fig:compare_tune_model_exp5}
\end{figure}

In this section, we perform a robustness analysis of our fine-tuning process. First, we examine how varying the level of noise added in (\ref{eq:lossmulti}) during estimation affects model performance. Specifically, we assess prediction accuracy on the test data by introducing noises with variance $\sigma^2 \in \{0, 0.001, 0.01, 0.05\}$. The resulting MSE under each noise setting is shown in Figure~\ref{fig:compare_tune_model_exp5}(a). The results indicate that the DSR model consistently outperforms the $L_2$ regression model across all noise levels. When $\sigma^2 = 0$, DSR and the $L_2$ regression model perform equivalently, whereas at $\sigma^2 = 0.05$, DSR achieves the best overall performance. Additionally, the 4DFlowNet model shows inferior performance compared to both DSR and the $L_2$ regression model under all noise conditions. These results suggest that adding a small amount of noise in DSR enhances super-resolution performance.

Next, we analyze whether the splitting of fine-tuning and testing datasets affects prediction accuracy. We sample different subsets of CFD and 4DF pairs for fine-tuning and use the remaining data as the testing dataset, repeating this procedure 30 times. Figure~\ref{fig:compare_tune_model_exp5}(b) presents box plots of the MSE across the 30 trials, showing that the DSR model consistently achieves lower MSE values than the $L_2$ regression model across different samplings. In contrast, the 4DFlowNet model exhibits the highest MSE among the three models in all trials, indicating poorer performance.

\graphicspath{{figs1/}}

\section{Conclusion, limitation and discussion}\label{sec:discussion}

We present a distributional learning framework specifically designed to enhance the extrapolation capabilities of super-resolution  methods. Additionally, we propose a comprehensive pipeline encompassing training, fine-tuning, and testing to improve the resolution of 4DF data on irregular 3D geometries. Our method is evaluated against non-distributional learning benchmarks, and experimental results demonstrate that the proposed framework significantly outperforms traditional super-resolution approaches, effectively enhancing the resolution of 4DF datasets.

Due to current data availability, the method has been evaluated on a limited set of test samples. We have collected 4DF data from approximately 150 patients over the past few years and are currently processing this dataset to support large-scale fine-tuning and evaluation. Once completed, this dataset will be made publicly available to facilitate further model development.

Since CFD simulations are based on physical principles, it is natural to consider that incorporating physics into the model could enhance estimation efficiency. However, our current model does not include physical constraints because it is designed to be geometry-invariant, while physical conditions are often closely tied to specific geometries. As a result, integrating physics into the framework would require significant additional development. We are actively exploring physics-informed deep learning in our ongoing research.

\section{Data Availability Statement}\label{data-availability-statement}

Deidentified data have been made available at the \hyperlink{https://github.com/wenxy123/DSR.git}{DSR}.


\bibliography{bibliography.bib}

\end{document}